\documentclass[preprint,12pt]{elsarticle}

\usepackage{amssymb}
\usepackage{amsthm}

\usepackage{siunitx}
\usepackage{subfigure}
\usepackage{times}
\usepackage{url}
\usepackage{epsfig}
\usepackage{caption}
\usepackage{graphicx}
\usepackage{amsmath,amssymb} 
\usepackage{mathtools}
\usepackage{booktabs}
\usepackage{amsfonts}
\usepackage{upgreek}
\usepackage[pagebackref=true,breaklinks=true,letterpaper=true,colorlinks,bookmarks=false]{hyperref}
\usepackage[usenames,dvipsnames,svgnames,table]{xcolor}

\usepackage{hyperref}  
\usepackage{capt-of}
\usepackage{dirtytalk}
\usepackage{algorithm}
\usepackage[noend]{algpseudocode}
\usepackage[super]{nth}
\usepackage[normalem]{ulem}

\usepackage{multirow}
\usepackage{subfigure}
\usepackage[table]{xcolor}
\usepackage{bm}
\usepackage{subfigure}
\usepackage{balance}
\usepackage{hhline,colortbl}
\usepackage{makecell}
\usepackage{comment}
\usepackage{fancyhdr} 
\usepackage{xcolor}
\definecolor{gray}{gray}{0.86}

\graphicspath{{./}{./figures/}}
\DeclareGraphicsExtensions{.pdf,.jpeg,.png,.jpg,.eps}

\def\etc{\emph{etc.}}
\def\eg{\emph{e.g.}}
\def\ie{\emph{i.e.}}

\newcommand{\RNum}[1]{\lowercase\expandafter{\romannumeral #1\relax}}

\newlength{\Oldarrayrulewidth}

\usepackage{float}

\newcommand{\printfnsymbol}[1]{%
  \textsuperscript{\textasteriskcentered}%
}

\journal{Computer Vision and Image Understanding}

\begin{document}
\begin{frontmatter}

\title{PointCaM: Cut-and-Mix for Open-Set Point Cloud Learning}

\author[1]{Jie Hong\fnref{fn1,fn2}}
\ead{jiehong@hku.hk}
\author[2]{Shi Qiu\fnref{fn1}}
\ead{shiqiu@cse.cuhk.edu.hk}
\author[3]{Weihao Li}
\ead{weihao.li1@anu.edu.au}
\author[3,4]{Saeed Anwar}
\ead{saeed.anwar@uwa.edu.au}
\author[5]{Mehrtash Harandi}
\ead{mehrtash.harandi@monash.edu}
\author[3]{Nick~Barnes}
\ead{nick.barnes@anu.edu.au}
\author[6]{Lars Petersson}
\ead{lars.petersson@data61.csiro.au}

\address[1]{The University of Hong Kong, Pok Fu Lam, Hong Kong SAR, China}
\address[2]{The Chinese University of Hong Kong, Shatin, Hong Kong SAR, China}
\address[3]{Australian National University, Canberra, Acton 2601, Australia}
\address[4]{The University of Western Australia, Perth, WA 6009, Australia}
\address[5]{Monash University, Wellington Rd, Clayton VIC 3800, Australia}
\address[6]{Data61-CSIRO, Canberra, Acton 2601, Australia}

\fntext[fn1]{Equal contributions.}
\fntext[fn2]{Corresponding author.}

\begin{abstract}
Point cloud learning is receiving increasing attention. However, most existing point cloud models lack the practical ability to deal with the unavoidable presence of unknown objects. This paper primarily discusses point cloud learning in open-set settings, where we train the model without data from unknown classes and identify them during the inference stage. In essence, we propose a novel Point Cut-and-Mix mechanism for solving open-set point cloud learning, comprising an Unknown-Point Simulator and an Unknown-Point Estimator module. Specifically, we use the Unknown-Point Simulator to simulate out-of-distribution data in the training stage by manipulating the geometric context of partially known data. Based on this, the Unknown-Point Estimator module learns to exploit the point cloud's feature context to discriminate between known and unknown data. Unlike existing methods that only consider classifier features, our proposed solution leverages multi-level feature contexts to recognize unknown point cloud objects more effectively. We test the proposed approach on several datasets, including customized S3DIS, ModelNet40, and ScanObjectNN. The improved open-set performances over comparative baselines show the effectiveness of our PointCaM method. Our code is available at \url{https://github.com/JHome1/pointcam}.
\end{abstract}


\end{frontmatter}

\section{Introduction}
To better illustrate complicated scenes in the real world, 3D data has been widely investigated in computer vision research. As a fundamental representation of 3D data, point clouds, which are data points typically in the 3D coordinate system and can be collected from 3D scanners~\cite{endres2013, jaboyedoff2012use}, have shown great potential in AI-related applications, such as robotics~\cite{pomerleau2015review, chen2019framework, ze2023visual, yu2024unified}, remote sensing~\cite{li2012new}, and augmented/virtual reality~\cite{chen2019overview, li2025dgns}.

Recent works~\cite{qi2017pointnet, qi2017pointnet++, zhao2021point, guo2020deep} leverage data-driven neural networks to analyze the context of 3D point clouds. Particularly, large-scale labeled point cloud data plays a crucial role in impacting the performance of deep networks. Despite the fast development in 3D sensing technology, point cloud data still requires expensive equipment for collection and huge human resources for annotation. The training datasets are assumed to contain all pre-defined object classes in the scenes; however, it is impractical to cover all possible classes in the real world. Thus, the richness and diversity of point cloud data are usually insufficient, especially for analyzing large-scale scenes~\cite{armeni2017joint, behley2019semantickitti} and a wide variety of objects~\cite{wu20153d, chang2015shapenet}. Moreover, the effectiveness of point cloud data is also limited by its inherent sparsity and irregularity, which cause further challenges in common visual tasks such as point cloud classification~\cite{qi2017pointnet, qiu2021geometric, li2020real}, segmentation~\cite{choy20194d, hu2020randla, qiu2021semantic}, and detection~\cite{qi2019deep, qiu2021investigating, li2023efficient}.

The state-of-the-art point cloud learning models~\cite{hu2020randla, wang2018sgpn, qiu2022pu} mainly focus on closed-set environments. These works often make a strong assumption that the object classes known during the deployment stage are only those provided during the training stage. However, this closed-set assumption is too restrictive, as autonomous systems will inevitably encounter unknown object classes that are not part of the training data in real-world environments. Currently, open-set recognition in the 2D image domain is an active area of research~\cite {scheirer2012toward, Bendale_2016_CVPR}, whereas the open-set learning of 3D point clouds is not commonly studied. In fact, this study has significant value in real-world applications. A possible outdoor scenario arises in self-driving, where an intelligent system may encounter objects that have not been seen before. By leveraging 3D open-set settings, the intelligent system can effectively identify such unknown objects, making drivers aware of the surroundings. Another potential use case for our 3D open-set settings is in XR (extended reality) systems. When a user is immersed in an XR gaming environment at home, the sudden appearance of a cat or a baby in the guarded area could pose a safety risk. However, traditional alerting mechanisms may not detect these unexpectedly moving creatures as their detection algorithms are trained on regular indoor furniture and objects under closed-set settings. By incorporating our proposed settings and solutions, the XR system can easily recognize such ``unknown'' creatures, thereby increasing the robustness and effectiveness of the alerting mechanism. The use cases of our study also lie in remote sensing applications: in urban planning, the proposed solution can be applied to detect unknown buildings, infrastructure, and land use changes from aerial LiDAR scans, allowing planners to better monitor city evolution; for environmental monitoring, the proposed solution can identify invasive plant species or track the impact of natural disasters using 3D sensing data, enabling more efficient allocation of rescue and recovery resources. Additionally, industrial automation systems can leverage our open-set point cloud learning to enhance both flexibility and safety. In reconfigurable production lines, the proposed approach can detect defects and anomalies in 3D scans of produced items, improving product quality control and simultaneously monitoring for unexpected objects to enhance workplace safety. Moreover, robotics in healthcare can benefit significantly from our open-set point cloud learning. When processing 3D medical scans, the proposed approach can identify unknown anatomical variations, aiding clinicians in diagnosis and treatment. In robotic-assisted surgery, the proposed method can help detect and recognize unexpected anatomical features, improving procedure precision and safety. Given the promises in academic research and real-world applications, we explore open-set point cloud learning by enhancing the models' abilities to deal with the inevitable unknown objects in 3D point cloud data.

\begin{figure}[t]
\begin{center}
\includegraphics[width=0.95\columnwidth]{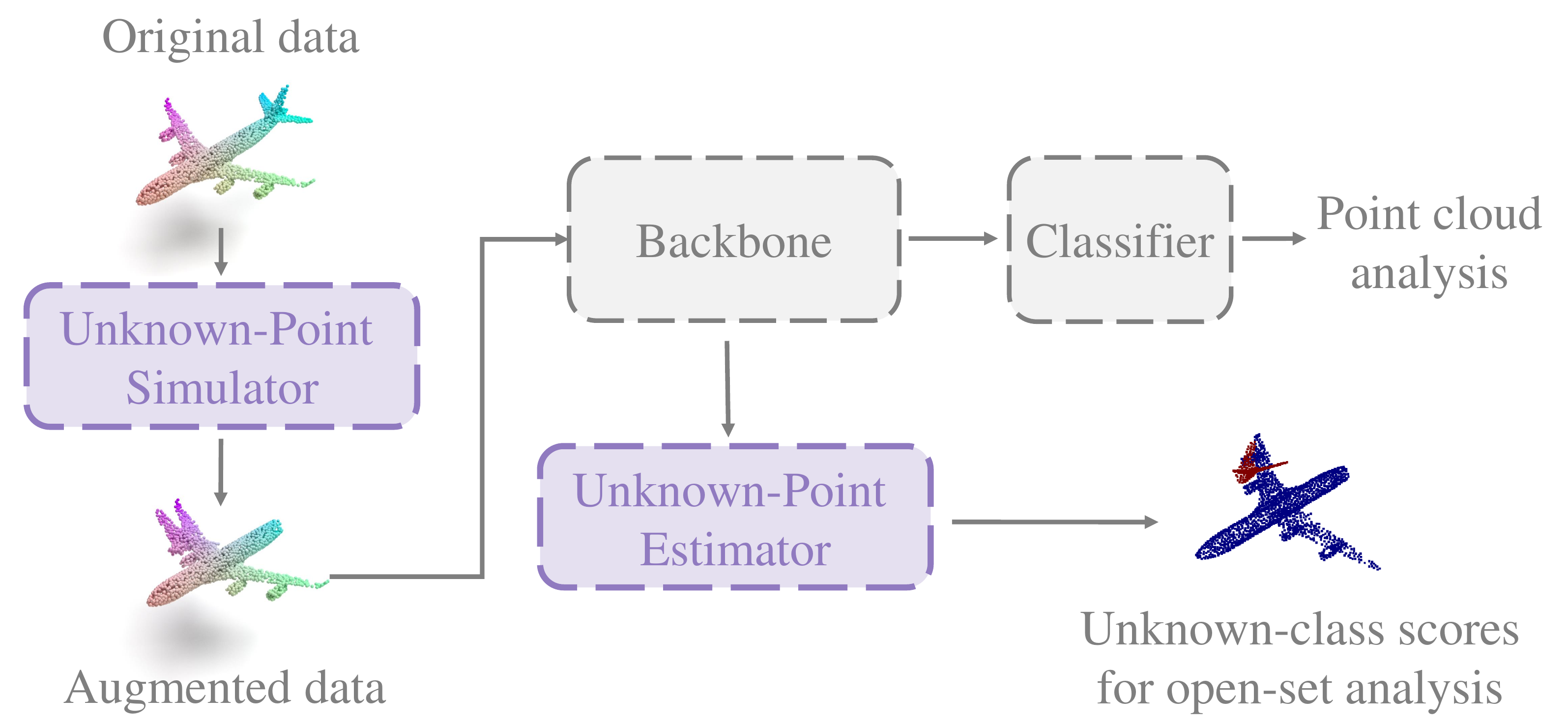}
\caption{The PointCaM mechanism consists of two main modules: the Unknown-Point Simulator (UPS) simulates the out-of-distribution data to some extent during the training, and the Unknown-Point Estimator (UPE) is responsible for computing the unknown-class scores.}
\label{fig:intro}
\end{center}
\vspace{-20pt}
\end{figure}

Recently, a few studies investigate 3D data under open-set settings \cite{ma2018towards, bhardwaj2021empowering}. For example, \cite{zhao2021few} addressed the few-shot 3D semantic segmentation task under an open-set setting, which presents an attention-aware multi-prototype transductive inference method. 
The 3D semantic segmentation task involves assigning a semantic label to each point. Transductive inference refers to the model's ability to access test samples without labels during the training stage.
The work~\cite{wong2020identifying} addresses unknown instance recognition in self-driving scenarios, where a feature predictor for anchors is designed to capture the characteristics of LiDAR data. Cen \emph{et al.}~\cite{cen2021open} study open-set LiDAR point cloud detection through metric learning, which is to learn the feature space using distance metrics to overcome the problem of massive false positive detection samples. Moreover, \cite{cen2022open} explores open-world LiDAR segmentation, where a redundancy classifier framework enables the model to adapt to the open-set setting and incremental learning tasks. The existing methods provide effective solutions to open-set 3D learning. However, these works have limitations that might affect the performance. First, methods~\cite{wong2020identifying, cen2021open, dong2022region, li2023open} lack simulated data about unknown classes, which might impact the model's sensitivity to unknown objects. Second, \cite{wong2020identifying, cen2021open, cen2022open, dong2022region, li2023open} only focus on the classifier's features for recognizing unknown objects while neglecting information from the data and learned feature maps.

To address the above limitations, in this work, we introduce a novel Point Cut-and-Mix (PointCaM) mechanism for open-set 3D point cloud learning. As illustrated in Fig.~\ref{fig:intro}, PointCaM incorporates two main modules: Unknown-Point Simulator (UPS)\footnote{UPS for segmentation and Unknown-Sample Simulator (USS) for classification.} and Unknown-Point Estimator (UPE)\footnote{UPE for segmentation and Unknown-Sample Estimator (USE) for classification. For simplicity, throughout the manuscript, we mainly use the terms: ``unknown point,'' ``UPS,'' and ``UPE.''}. During the training process, we leverage the Unknown-Point Simulator to simulate out-of-distribution data to some extent by partially manipulating the geometric context of point clouds. To further discriminate the known and unknown objects provided by the UPS module, our UPE is designed to calculate an unknown-class score for each point. In particular, the UPE module estimates the unknown-class scores by adaptively fusing the multi-level feature context of point cloud data to learn the geometric dependencies and semantic correlation between known and unknown data.

The main contributions of this paper are summarized as follows: 
(1) We design a novel mechanism, PointCaM, for open-set point cloud learning. Two-component modules, the Unknown-Point Simulator (UPS) and the Unknown-Point Estimator (UPE), are proposed to simulate out-of-distribution data and estimate the unknown-class scores, respectively. 
(2) UPS helps to simulate out-of-distribution objects, where the selected point clouds are rotated and translated to exhibit highly distinct geometric distributions from known objects. UPS is the first to employ Cut-Mix augmentation, providing a more distinct geometric distribution of simulated data than prior works.
(3) UPE adaptively fuses information from multi-level feature contexts for computing the unknown scores, which addresses the narrow focus of the earlier studies on classifier features. By utilizing information from both 3D coordinates and feature map representations, UPE can leverage rich geometric and semantic contexts for differentiating between known and unknown objects.
\section{Related Work}
\subsection{Open-set Recognition}
Although modern deep learning has demonstrated great success in closed-set visual recognition tasks, the challenge of open-set recognition has gained attention~\cite{scheirer2012toward, Bendale_2016_CVPR}. In a closed-set image recognition scenario, the classifier aims to identify a set of image categories that remain consistent during both the training and testing phases. On the other hand, in an open-set image recognition scenario, the classifier must differentiate between the training categories and determine if an image belongs to an unknown class. In the case of object detection, open-set detection~\cite{dhamija2020overlooked, joseph2021towards, Zheng_2022_CVPR} is designed to detect unknown objects from open-set data that includes both known and unknown objects. Similarly, in open-set semantic segmentation~\cite{hendrycks2019scaling, hong2022goss, hong2023curved}, the goal is to segment an image containing known and unknown object classes. In addition to the task settings above, researchers have also been actively exploring solutions in open-set learning. Recent studies have proposed methods such as density estimation~\cite{zhang2020hybrid}, uncertainty modeling~\cite{liang2017enhancing}, and input reconstruction~\cite{pidhorskyi2018generative} to identify open-set examples. For instance, \cite{zhang2020hybrid} uses an encoder to encode the input data into a joint embedding space, a classifier to classify samples into known classes, and a flow-based density estimator to detect whether a sample belongs to the unknown class. \cite{pidhorskyi2018generative} introduces a network architecture for open-set recognition based on learning reconstruction of input data and detects unknown samples by evaluating its inlier probability distribution. 
Nonetheless, all these studies still primarily concentrate on 2D recognition tasks and datasets, and open-set learning of 3D point clouds is rarely studied. In this work, we introduce a new 3D open-set setting for the point cloud domain.

\subsection{Point Cloud Learning}
Following the success of Convolutional Neural Networks (CNNs) in image recognition, several papers~\cite{guo2020deep} have developed dedicated CNN-based models to analyze point cloud data. In general, point cloud learning relies highly on the learned point feature representations for accurate visual recognition. Recent networks can be categorized into two main streams: indirect and direct methods. 
Specifically, the indirect methods project original point clouds into certain intermediate representations, \eg, multi-view images~\cite{su2015multi} or voxels~\cite{maturana2015voxnet}, and apply regular 2D/3D CNNs to analyze the projected data. More recently, researchers tend to directly analyze given point cloud data using the tools of multi-layer perceptron (MLP)~\cite{qi2017pointnet,qi2017pointnet++}, graph structure~\cite{wang2019dynamic}, transformer~\cite{zhao2021point, wu2022point, wu2024point}, \etc~Compared to indirect methods, direct methods can save complicated data processing, avoid information loss, and benefit practical feasibility. In this work, we focus on direct open-set point cloud learning, including the basic visual tasks of semantic segmentation and classification. 

\subsection{3D Open-set Settings}
Recently, a few works have emerged addressing 3D data under open-set settings~\cite{ma2018towards, bhardwaj2021empowering}. 
As the first work doing few-shot 3D learning, the few-shot 3D semantic segmentation task is studied in \cite{zhao2021few}, where an attention-aware multi-prototype transductive inference method is proposed. Afterward, more solutions are studied for few-shot 3D learning~\cite{mao2022bidirectional, he2023prototype, xu2023generalized, an2024rethinking}. Different from open-set 3D learning, which recognizes the unknown point cloud objects from data, few-shot 3D learning focuses on classifying unknown point cloud objects given a limited number of labeled examples.
In \cite{wong2020identifying}, a feature predictor for anchors is specially designed to fit the characteristics of LiDAR data, targeting unknown instance recognition under self-driving scenarios. Cen \emph{et al.}~\cite{cen2021open} study open-set point cloud detection via metric learning to overcome the problem of massive false positive detection samples in LiDAR-based data. Moreover, open-world LiDAR segmentation is explored in \cite{cen2022open}, where a redundancy classifier framework is used to adapt the model to the open-set setting and incremental learning tasks. 
However, prior studies might have the following limitations: (1) methods in \cite{wong2020identifying, cen2021open, dong2022region, li2023open} lack simulated data regarding unknown classes, which may impact the model's sensitivity to unknown objects; and (2) \cite{wong2020identifying, cen2021open, cen2022open, dong2022region, li2023open} only focus on the classifier's features for recognizing unknown objects, while neglecting information from the data and learned feature maps. To address these limitations, our approach, PointCaM, utilizes parts of the point cloud to approximate the unknown data roughly. Specifically, we rotate and translate these point cloud sections to ensure sufficient geometric pattern differences from known objects. We then leverage the contextual information of the point cloud in both geometric and feature spaces to identify unknown data. Particularly, we adaptively fuse information from feature spaces with guidance from geometric space to compute the unknown score. The rich information concerning differences between known and unknown objects in such spaces enables the model to differentiate between the known and unknown categories more effectively.

\subsection{Cut-Mix Augmentation}
Cut-Mix and its related data augmentations offer a straightforward approach to enhancing visual recognition performance, including object detection~\cite{Ghiasi_2021_CVPR}, self-supervised representation learning~\cite{han2022you}, image classification~\cite{Yun_2019_ICCV}, and anomaly detection~\cite{Li_2021_CVPR}. 
For example, Cut-Paste-and-Learn~\cite{Dwibedi_2017_ICCV} proposes cutting foreground object instances and pasting them onto diverse background images to synthesize object detection training images with bounding box labels.
Copy-Paste~\cite{Ghiasi_2021_CVPR} demonstrates that applying random scale jittering to two randomly selected training images can significantly benefit training instance segmentation models.
CutMix~\cite{Yun_2019_ICCV} cuts a rectangular image patch from an image and pastes it at a random location of another image.
CutPaste~\cite{Li_2021_CVPR} cuts an image patch and pastes it at a random location of the same image, producing spatial irregularities to estimate real defects.
In open-set recognition, data augmentations are used to generate samples that simulate the presence of out-of-distribution classes during training. This can help the model learn more robust features that are less specific to the known classes and more generalizable to new or unseen classes. 
For example, \cite{neal2018open} introduces counterfactual image generation augmentation, which generates examples that are close to training set examples yet do not belong to any of the training categories, using generative adversarial networks.
Unlike the existing works mentioned above, we present the Cut-and-Mix mechanism targeting open-set point cloud learning. Our proposed mechanism is the first to simulate out-of-distribution point cloud data by rotating and translating existing points.
\section{Problem Statement}
\subsection{Open-set Point Cloud Learning}
Open-set point cloud learning includes two tasks: open-set point cloud segmentation and classification. The objective of open-set point cloud segmentation (or classification) is to assign an unknown score to each point (or sample). The deep model is trained solely on data from $N_c$ known classes, but during evaluation, it must handle both $N_c$ known and unknown classes. Assuming there is one unknown class, we calculate an unknown-class score, $x$ (or $x_{cls}$), for each testing point (or sample) to estimate its probability of belonging to the unknown class in segmentation (or classification). The overall performance of open-set point cloud learning is measured based on the unknown-class scores of all testing points (or samples), denoted as $\{x_1, x_2, ..., x_i, ...\}$ (or $\{x_{cls,1}, x_{cls,2}, ..., x_{cls,j}, ...\}$). Ideally, if the $i$-th point (or $j$-th sample) belongs to the unknown class, we hope that $x_i$ (or $x_{cls,j}$) approaches $1$; while the point (or sample) belongs to the known classes, the corresponding score is expected to be $0$. In summary, the goal of open-set point cloud learning is to train a network that can generate accurate unknown scores for each point (or sample).

\subsection{Benchmarks}
\label{sec:benchmarks}
We adopt three widely used point cloud learning datasets, S3DIS \cite{armeni2017joint}, ModelNet40 \cite{wu20153d}, and ScanObjectNN \cite{uy2019revisiting} to conduct our experiments on the proposed 3D open-set tasks. Specifically, the S3DIS~\cite{armeni2017joint} dataset contains $6$ large-scale workspace point cloud scans, which can be further divided into $272$ different rooms, including lobbies, offices, hallways, and conference rooms. In each room, about $0.5$ to $2.5$ million points are labeled into $13$ classes for semantic segmentation.
ModelNet40~\cite{wu20153d} consists of $12,311$ synthetic 3D object meshes in $40$ different categories, where the corresponding point cloud data is sampled from the surface of each object mesh via uniform sampling. Following the common practice in the previous classification works~\cite{qi2017pointnet,qi2017pointnet++}, about $9,843$ point clouds are used for training, while the remaining $2,468$ samples are reserved for testing. As a real-world counterpart to ModelNet40, the ScanObjectNN~\cite{uy2019revisiting} dataset comprises a total of $14,298$ manually collected point clouds across $15$ object categories. According to the official data split in~\cite{uy2019revisiting}, the training set comprises $11,416$ samples, while the remaining $2,882$ samples are reserved for testing purposes. To better evaluate our approaches, we conduct the open-set experiments under a more challenging case using the hardest perturbation variant of ScanObjectNN data~\cite{uy2019revisiting}.

In this open-set study, we customize each dataset by manually dividing the original data into known and unknown classes based on the given semantic labels. 
For open-set point cloud semantic segmentation, we borrow the idea of splitting classes in \cite{cen2021deep} where a few ``thing'' classes are chosen as unknown classes.
Practically, in the ``Manual-10-3'' split of S3DIS, we choose three ``thing'' classes of ``table,'' ``chair,'' and ``sofa'' to form the unknown class, while in the ``Manual-12-1'' split, only ``sofa'' is selected as the unknown class. For open-set point cloud classification, we divide each original dataset (ModelNet40~\cite{wu20153d} or ScanObjectNN~\cite{uy2019revisiting}) into two splits, ModelNet40\textit{-Split1} (or Scan-ObjectNN\textit{-Split1}) and ModelNet40\textit{-Split2} (or ScanObjectNN\textit{-Split2}). In particular, we purposely separate classes with similar appearances in different splits to increase the task difficulty. Suppose the data chosen as the unknown class has attributes similar to the data pre-defined as the known classes, and thus, it is more challenging for the models to distinguish between the unknown and the known classes. More details of the splits are provided in the supplementary material. 

\subsection{Evaluations}
Both closed-set and open-set metrics are evaluated in our experiments. We choose mIoU and classification accuracy as the closed-set metrics. The standard open-set metrics for 2D tasks are included as open-set metrics: the false positive rate at 95\% true positive rate (FPR at 95\% TPR), the area under the receiver operating characteristics (AUROC) \cite{davis2006relationship}, and the area under the precision-recall (AUPR) \cite{manning1999foundations}. Such metrics assess the performance based on the overlap of unknown-class score distributions between the known and unknown classes. We aim to enhance the models' open-set performance while preserving their closed-set classification accuracy. In other words, the model is expected to identify unknown objects and correctly classify known ones.
\section{Methodology}
\subsection{Baseline Method}
We adopt the training process of regular point cloud learning in the baseline models. Regarding the testing process, we borrow ideas from 2D tasks to compute the scores for unknown classes. Specifically, we employ Maximum Softmax Probability (MSP) \cite{hendrycks2016baseline} or Maximum Unnormalized Logit (MaxLogit) \cite{hendrycks2019scaling} to calculate an unknown-class score $x_i$ (or $x_{cls,j}$) for each point (or sample), based on its corresponding confidence or logit vector in the classifier.

\subsection{PointCaM Mechanism}
Given a point cloud containing $N$ points, we describe the 3D coordinates $\mathcal{P}\in\mathbb{R}^{N\times3}$ as the geometric context, which is explicitly collected by scanners, indicating the geometric distribution of points in the original 3D space. In addition, the feature context $\mathcal{F}\in\mathbb{R}^{N\times C}$ of point cloud can be implicitly learned via CNN-based operations/modules from both local and global perspectives~\cite{yan2020pointasnl,qiu2021pnp}, representing the latent semantic information in a $C$-dimensional feature space. In general, the geometric context $\mathcal{P}$ and the feature context $\mathcal{F}$ are considered two main properties of point cloud data~\cite{hu2020randla,qiu2021geometric}, which benefit comprehensive point feature representations for accurate visual analysis. Based on these two properties, we propose a novel self-supervised approach, the PointCaM mechanism, to enhance open-set point cloud learning.

The PointCaM mechanism is illustrated in Fig.~\ref{fig:intro}. Firstly, the Unknown-Point Simulator (UPS) or USS is applied to manipulate the geometric context partially, simulating unknown objects to some extent for training purposes. Secondly, a CNN-based Unknown-Point Estimator (UPE) or USE learns to discriminate between known and unknown objects by heavily exploiting feature context. Compared to the baseline method mentioned above, our PointCaM not only simulates out-of-distribution data for improved training perception but also extracts more semantically meaningful feature information from the intermediate encoders, rather than the output classifiers.

\subsubsection{Unknown-Point Simulator} 
Our main concern lies in simulating out-of-distribution objects while only known data are available during the training process. As mentioned above, the geometric context of point cloud data indicates the original distribution of known objects in 3D space. If we alter the geometric context to a certain extent, part of the known data will become unfamiliar to a CNN's perception, as the distribution of points partially changes. 
Using this approach, we propose Alg.~\ref{alg:aug} to generate a partial geometric context. 

\begin{algorithm}[h]
\caption{Unknown-Point Simulator (UPS)}\label{alg:aug}
\textbf{input:} a point cloud $\mathcal{P}\in\mathbb{R}^{N\times3}$\\
\textbf{output:} an augmented point cloud $\mathcal{Q}\in\mathbb{R}^{N\times3}$\\
{{\fontfamily{qcr}\selectfont\textcolor{Gray}{\# Randomly select a seed point $s$}}}\\
1. $s$ = \textbf{Rand}($\mathcal{P}$) $\in\mathbb{R}^{3}$\\
{{\fontfamily{qcr}\selectfont\textcolor{Gray}{\# Find the k nearest neighbors of $s$}}}\\
2. $\mathcal{C}$ = \textbf{Knn}($s$, $\mathcal{P}$) $\in\mathbb{R}^{k\times 3}$\\
{{\fontfamily{qcr}\selectfont\textcolor{Gray}{\# Augment $\mathcal{C}$ by rotation and translation}}}\\
3. $\mathcal{C}^\prime$ = \textbf{Aug}($\mathcal{C}$) = $T$ + $R \cdot {\mathcal{C}}$ $\in\mathbb{R}^{k\times 3}$\\
{{\fontfamily{qcr}\selectfont\textcolor{Gray}{\# Cut $\mathcal{C}$ from $\mathcal{P}$, and mix with $\mathcal{C}^\prime$}}}\\
4. $\mathcal{Q}$ = \textbf{CutMix}($\mathcal{C}$, $\mathcal{P}$, $\mathcal{C}^\prime$) $\in\mathbb{R}^{N\times3}$
\end{algorithm}

At first, we randomly select (``\textbf{Rand}'') a point $s\in\mathbb{R}^{3}$ from original point cloud $\mathcal{P}\in\mathbb{R}^{N\times3}$ as a seed point.  
{Randomly choosing the seed point from the entire set of point clouds ensures that UPS can simulate objects with a variety of appearances.}
Then, we search for the neighbors of $s$ as a subset of points $\mathcal{C}\in\mathbb{R}^{k\times3}$ that we aim to augment. In practice, we apply the k-nearest-neighbors (``\textbf{Knn}'') algorithm~\cite{wang2019dynamic} to find a set of $k$ neighbors based on the point-wise Euclidean distances in 3D space. Particularly, the number of neighbors follows $k = \beta \cdot N$, where the selection ratio $\beta$ is randomly drawn from $[\beta_{min}, \beta_{max}]$ for a point cloud of $N$ points. 
For neighbor selection, we define a specific range within which a subset of point clouds is randomly chosen, ensuring that simulated objects can appear at different scales.
An ablation study about the maximum selection ratio $\beta_{max} \in (0, 1)$ is provided in Tab.~\ref{table:beta}.
{Another ablation study in training time, which is shown in Tab.~\ref{table:complex}, demonstrates that the point selection of UPS is not costly compared to the baseline.}
To augment (``\textbf{Aug}'') $\mathcal{C}$ as unknown points, we incorporate basic rotation and translation operations, considering two main benefits: (1) Compared to adding noise or scaling, it is a better way of simulating unknown objects, as the natural local structures are preserved. This is evidenced by our experimental results presented in Tab.~\ref{table:generator}, where different regular transformations are performed to simulate unknown data points. Notably, we observe that incorporating random rotation and translation can improve performance. The scaling operation has a trivial effect, whereas adding Gaussian noise inevitably damages the local structures of an object, negatively impacting its performance. 
(2) Since only a subset of points ($k \ll N$) are modified with regular transformations. In contrast, a majority of points remain unchanged; we ensure that for each point cloud sample, a proper proportion exists between known and unknown data points. This process is important to the 3D open-set settings: on the one hand, a small number of transformed points (\ie, the ones simulated as ``out-of-distribution data'') can help the model develop an additional capability of differentiating open-set data; on the other hand, a large number of unchanged points (\ie, the ones used as ``known data'') can help the model maintain its original capability of recognizing close-set data for regular 3D semantic segmentation and classification tasks.
Specifically, $R\in\mathbb{R}^{3\times3}$ in Alg.~\ref{alg:aug} is a general rotation matrix representing three degree-of-freedom (3-DoF) random rotations, and $T\in\mathbb{R}^{3}$ is a 3-DoF random vector representing a further translation within the borders of $\mathcal{P}$. 
Finally, we drop the partial points $\mathcal{C}$ in the original $\mathcal{P}$ and keep the transformed $\mathcal{C}^\prime$ as the simulated {out-of-distribution} points,
obtaining the partially altered geometric context $\mathcal{Q}\in\mathbb{R}^{N\times3}$ that contains both known and unknown points (``\textbf{CutMix}''). 

Using the UPS module, the network is still trained with a standard segmentation or classification loss, \ie, $\ell = \ell_{task}$.
In open-set point cloud semantic segmentation, the selected points are assigned a label of $N_c+1$, where $N_c$ is the number of known classes. 
{Similarly, in open-set point cloud classification, two samples of different known classes are partially cut, transformed, and mixed to form a new sample labeled as $N_c+1$, which introduces an additional class for unknown data in addition to the $N_c$ known classes during training. The goal of UPS is to enrich the model’s knowledge of unknown objects to some degree, enhancing its sensitivity to such data.}
To achieve this, we generate unknown points in the data by manipulating parts of the original point cloud to ensure that the simulated points have reasonable geometric features. Additionally, the rotation and translation of these points exhibit an abnormal geometric distribution from the known points.
Notably, since the UPS module simulates out-of-distribution data as a new class for training, one more dimension is added to the classifier's output accordingly.

\subsubsection{Unknown-Point Estimator}
\begin{figure}
\begin{center}
\includegraphics[width=0.98\columnwidth]{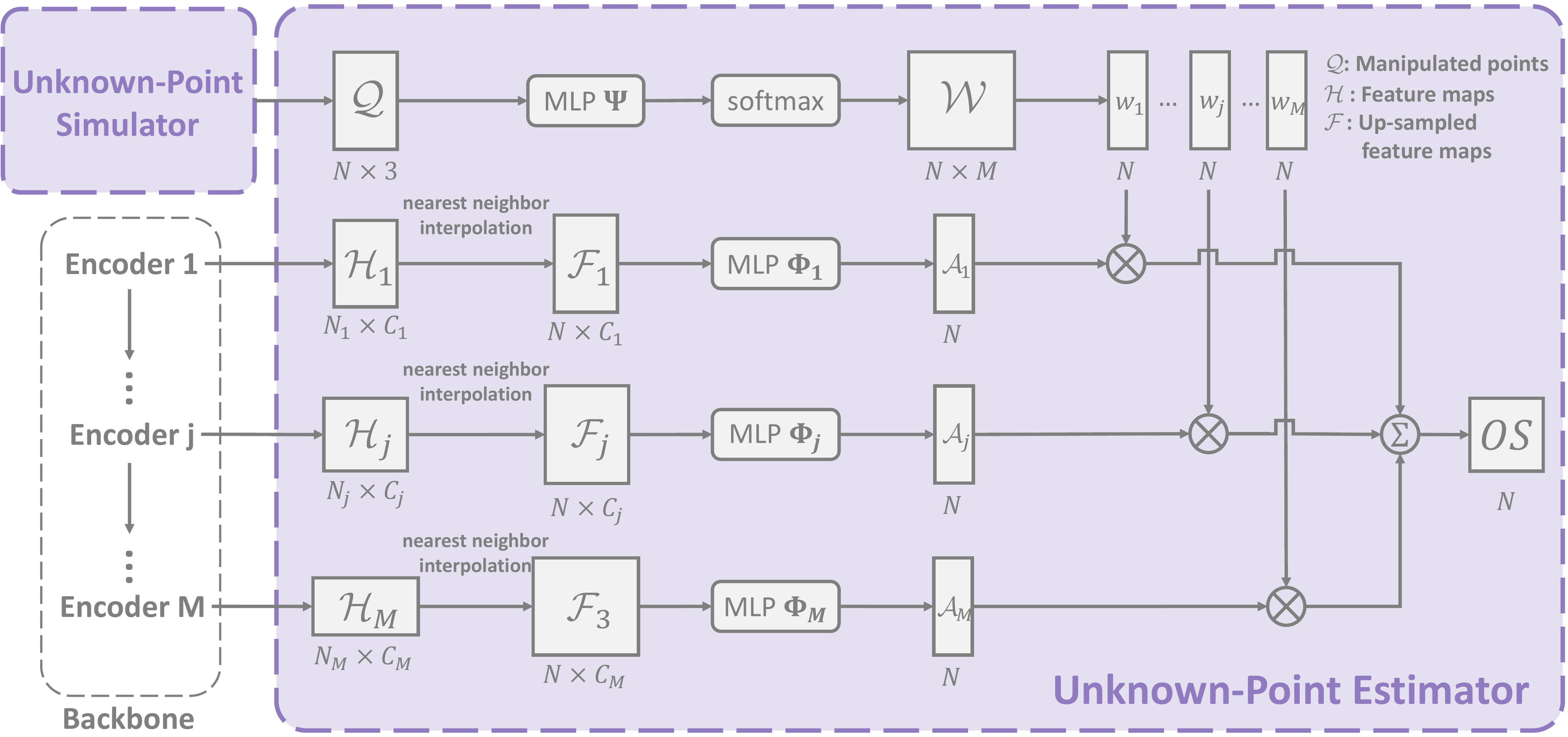}
\caption{The detailed structure of the Unknown-Point Estimator (UPE). The figure illustrates the operations of the UPE module during the training process. We only demonstrate an example of fusing three intermediate feature maps for simplicity. $j$ is an arbitrary integer satisfying $1\leq j \leq M$.}
\label{fig:upe}
\end{center}
\vspace{-20pt}
\end{figure}

To discriminate the known and unknown points in $\mathcal{Q}$, a CNN-based Unknown-Point Estimator (UPE) module learns to compute accurate unknown-class scores by exploiting the feature context in the point cloud. As depicted in Fig.~\ref{fig:upe}, given a backbone network, we denote the intermediate feature maps extracted from the backbone's encoders as $\{\mathcal{H}_1, ..., \mathcal{H}_j, ..., \mathcal{H}_M\}$, where $M$ is the number of intermediate feature maps and $\mathcal{H}_j$ stands for an arbitrary one satisfying $1\leq j \leq M$. Since a few point cloud networks~\cite{qi2017pointnet++,choy20194d,zhao2021point} follow a U-Net~\cite{ronneberger2015u} architecture to learn multi-level feature maps from lower point cloud resolutions (\ie, $\mathcal{H}_j\in\mathbb{R}^{N_j\times C_j}$, where $N_j < N$), the extracted feature maps usually have different sizes. To restore the full feature context for all $N$ points, we upsample $\{\mathcal{H}_1, ..., \mathcal{H}_j, ..., \mathcal{H}_M\}$ via the nearest neighbor interpolation method~\cite{hu2020randla}. Then, the corresponding full-size feature context representations are obtained as $\{\mathcal{F}_1, ..., \mathcal{F}_j, ..., \mathcal{F}_M\}$, where $\forall\mathcal{F}_j\in\mathbb{R}^{N\times C_j}$. 

Based on the feature context of $\{\mathcal{F}_1, ..., \mathcal{F}_j, ..., \mathcal{F}_M\}$, we intend to squeeze the latent information and estimate a score $x_i$ evaluating each point's probability
of whether it belongs to the unknown data points. To achieve this, we apply a set of regular MLPs $\{\bm{\Phi_1}, ..., \bm{\Phi_j}, ..., \bm{\Phi_M}\}$ to regress $\{\mathcal{F}_1, ..., \mathcal{F}_j, ..., \mathcal{F}_M\}$, respectively. For an arbitrary feature context $\mathcal{F}_j$, the estimated score vector is computed as $\mathcal{A}_j = \bm{\Phi_j}(\mathcal{F}_j)$. Following a similar computation, we acquire a set of score vectors $\{\mathcal{A}_1, ..., \mathcal{A}_j, ..., \mathcal{A}_M\}$, where $\forall \mathcal{A}_j \in\mathbb{R}^{N}$. Since the estimated score vectors are derived from different scales/levels of feature context representing a network's hierarchical semantic information, it is necessary to synthesize them for a comprehensive prediction. Particularly, we leverage the altered geometric context $\mathcal{Q}$ to guide an adaptive fusion of $\{\mathcal{A}_1, ..., \mathcal{A}_j, ..., \mathcal{A}_M\}$, because the simulated {out-of-distribution} data points are implied by their anomalous geometric distributions in 3D space compared to those given (known) points. To be concrete, we utilize another MLP $\bm{\Psi}$ followed by a channel-wise softmax function to learn the weight vectors:
\begin{equation}
    \mathcal{W} = [w_1; ...; w_j; ...; w_M] = \mathrm{softmax}\big(\bm{\Psi}(\mathcal{Q})\big),
\end{equation}
where $\mathcal{W}\in\mathbb{R}^{N\times M}$, and $\forall w_j\in\mathbb{R}^{N}$ represents a weight vector split from the columns of $\mathcal{W}$. Finally, we fuse the score vectors for all points as a weighted sum:
\begin{equation}
\label{eq:X}
    \mathcal{X} = \sum_{j=1}^{M}{w_j \times \mathcal{A}_j}, \quad \mathcal{X} \in\mathbb{R}^{N}.
\end{equation}
The output vector $\mathcal{X}$'s $i$-th scalar, $x_i$, corresponds to our computed unknown-class score on the $i$-th point $q_i$ of $\mathcal{Q}$, indicating its possibility of being a simulated unknown data point.  
When training the UPE module, in addition to a typical task loss $\ell_{task}$, the model is also subject to a mean squared error (MSE) loss $\ell_{upe}$ computed upon the UPE's output score vector $\mathcal{X}$. The computation of $\ell_{upe}$ is formulated as:
\begin{equation}
    \ell_{upe} = \frac{1}{N}\sum_{i=1}^{N} {(x_i - \hat{x_i})}^2;
\end{equation}
where the reference score $\hat{x_i} = 1$ if the point $q_i$ is a simulated {out-of-distribution} data point; and $\hat{x_i} = 0$ if the point $q_i$ belongs to one of the known classes. 
By minimizing the total loss $\ell = \ell_{task} + \alpha \cdot \ell_{upe}$ ($\alpha$ is an empirical parameter studied in Tab.~\ref{table:alpha}) during the training stage, we enable the UPE module to differentiate between the unknown and known data points. In the testing stage, we directly calculate the unknown-class scores $\mathcal{X}$ following Eq.~(\ref{eq:X}), where the $i$-th scalar value $x_i$ predicts the possibility of $\mathcal{Q}$'s $i$-th point $q_i$ belonging to the unknown class. 
For the semantic segmentation task, unknown scores are calculated for all points to facilitate prediction. The classification task extracts an averaged unknown score over $N$ points of the given sample for prediction.

UPE enhances the open-set metric by investigating geometric dependencies and semantic connections between known and unknown points. To accomplish this, UPE employs an adaptive multi-level fusion module to distinguish between known and artificially generated unknown points. By examining point clouds and feature maps, UPE obtains more detailed information about the distinctions between known and unknown points.
Compared to traditional multi-level prediction models~\cite{ronneberger2015u, long2015fully, lee2015deeply}, which only employ feature maps, our UPE leverages the 3D geometric context of the point cloud to compute weights for the adaptive multi-level fusion. This enables the adaptive and effective integration of multi-level semantic information in feature space, guided by point cloud geometric structures in 3D space.
\section{Experiments}\label{Experiment}
Experiments have been conducted on two fundamental open-set point cloud learning tasks: point cloud semantic segmentation and classification. We test methods on customized point cloud datasets: S3DIS\textit{-Split} for open-set 3D semantic segmentation, and ModelNet40\textit{-Split} and ScanObjectNN\textit{-Split} for open-set 3D classification. In general, we present rich experimental data to indicate the feasibility of our open-set settings for point cloud learning. Coupled with more ablation studies, our PointCaM mechanism is comprehensively verified using different benchmarks and backbone networks.

\subsection{Implementations}
In open-set 3D semantic segmentation experiments, $k$ points are rotated and translated to mimic unknown points. Note $k = \beta \cdot N$ where $\beta \in [\beta_{min}, \beta_{max}]$ is the random selection ratio, and $N$ is the total number of points of the point cloud. PointNet and PointNet++ are trained with $64$ epochs using Adam optimizer~\cite{kingma2014adam}. The initial learning rate is set to $0.001$. PointTransformers are trained with $100$ epochs using the SGD optimizer. The initial learning rate is $0.5$. For open-set 3D classification, we first randomly select samples to be augmented at an augmentation ratio of $0.1$; then, for each chosen sample, we perform the augmentation process by mixing it with $k$ points randomly selected from another sample. Specifically, a total of $k$ points can be selected from another sample of $N$ points $(k = \beta \cdot N)$. Then, we augment and mix these $k$ neighbor points with the chosen sample points. A mixed sample will be used as a sample from the unknown class. All classification models are trained with $300$ epochs using an SGD optimizer. The number of total points, \ie, $N$, is set to $4096$ and $1024$ for open-set point cloud semantic segmentation and classification, respectively. We run all experiments using the NVIDIA GeForce RTX 3090 GPU.

\begin{table*}[h]
\begin{center}

\resizebox{.69\textwidth}{!}{
\begin{tabular}{l|l|l|c|ccc}

\hline
&& &\textbf{Closed-Set Metric} &\multicolumn{3}{c}{\textbf{Open-set Metric}}
\\ \hline

\textbf{Data Split} &\textbf{Backbone} &\textbf{Open-set Method} &\textbf{mIoU} &\textbf{FPR (95\% TPR) $\downarrow$} &\textbf{AUPR $\uparrow$} &\textbf{AUROC $\uparrow$} 
\\ \hline

\multirow{28}*{Manual-12-1}
&\multirow{7}*{\makecell[l]{PointNet~\cite{qi2017pointnet}}} 
                                &MSP~\cite{hendrycks2016baseline}       &48.8 &47.1 &0.9 &81.3 \\
&                               &MaxLogits~\cite{hendrycks2019scaling}  &48.8 &45.5 &1.3 &82.3 \\
&                               &REAL~\cite{cen2022open} &46.5 &44.7 &0.9 &81.8 \\
&                               &UPS+MSP         &47.1 &47.4 &1.1 &80.5 \\
&                               &UPS+MaxLogits   &47.1 &45.3 &1.0 &81.4 \\
&                               &\cellcolor{gray}  &\cellcolor{gray} &\cellcolor{gray} &\cellcolor{gray} &\cellcolor{gray} \\
&                               &\cellcolor{gray}\multirow{-2}*{\makecell[l]{PointCaM \\ (UPS+UPE, Ours)}} &\cellcolor{gray}\multirow{-2}*{46.5} &\cellcolor{gray}\multirow{-2}*{40.9} &\cellcolor{gray}\multirow{-2}*{0.9} &\cellcolor{gray}\multirow{-2}*{\textbf{83.1}} \\ \cline{2-7}

&\multirow{7}*{\makecell[l]{PointNet++~\cite{qi2017pointnet++}}}
                                &MSP~\cite{hendrycks2016baseline}      &52.8 &93.4 &0.3 &51.3 \\
&                               &MaxLogits~\cite{hendrycks2019scaling} &52.8 &92.7 &0.3 &48.8 \\
&                               &REAL~\cite{cen2022open} &53.7 &93.1 &0.3 &51.9 \\
&                               &UPS+MSP        &53.7 &93.1 &0.3 &52.4 \\
&                               &UPS+MaxLogits  &53.7 &93.4 &0.3 &50.8 \\
&                               &\cellcolor{gray}  &\cellcolor{gray} &\cellcolor{gray} &\cellcolor{gray} &\cellcolor{gray} \\
&                               &\cellcolor{gray}\multirow{-2}*{\makecell[l]{PointCaM \\ (UPS+UPE, Ours)}}
&\cellcolor{gray}\multirow{-2}*{54.8} &\cellcolor{gray}\multirow{-2}*{87.4} &\cellcolor{gray}\multirow{-2}*{0.4} &\cellcolor{gray}\multirow{-2}*{\textbf{55.6}} \\ \cline{2-7}

&\multirow{7}*{\makecell[l]{PointTransformer~\cite{zhao2021point}}} 
                                &MSP~\cite{hendrycks2016baseline}      &66.5 &100.0 &13.3 &59.0 \\
&                               &MaxLogits~\cite{hendrycks2019scaling} &66.5 &99.8  &13.4 &59.3 \\
&                               &REAL~\cite{cen2022open} &67.5 &99.8 &14.4 &62.0 \\
&                               &UPS+MSP        &67.1 &99.7 &17.0 &66.0 \\
&                               &UPS+MaxLogits  &67.1 &90.1 &18.2 &65.4 \\
&                               &\cellcolor{gray}  &\cellcolor{gray} &\cellcolor{gray} &\cellcolor{gray} &\cellcolor{gray} \\
&                               &\cellcolor{gray}\multirow{-2}*{\makecell[l]{PointCaM \\ (UPS+UPE, Ours)}}
&\cellcolor{gray}\multirow{-2}*{67.1} &\cellcolor{gray}\multirow{-2}*{73.9} &\cellcolor{gray}\multirow{-2}*{18.7} &\cellcolor{gray}\multirow{-2}*{\textbf{72.0}} \\ \cline{2-7}

&\multirow{6}*{\makecell[l]{PointTransformerV3~\cite{wu2024point}}}
                                &MSP~\cite{hendrycks2016baseline}       &66.0 &100.0 &13.6 &59.3 \\
                                
&                               &MaxLogits~\cite{hendrycks2019scaling}  &66.0 &90.8	&16.2 &63.1 \\
&                               &UPS+MSP         &64.1 &99.9 &19.5 &68.3 \\
&                               &UPS+MaxLogits   &64.1 &85.3 &20.4 &69.5 \\
&                               &\cellcolor{gray}  &\cellcolor{gray} &\cellcolor{gray} &\cellcolor{gray} &\cellcolor{gray} \\
&                               &\cellcolor{gray}\multirow{-2}*{\makecell[l]{PointCaM \\ (UPS+UPE, Ours)}} 
&\cellcolor{gray}\multirow{-2}*{64.2} &\cellcolor{gray}\multirow{-2}*{59.4} &\cellcolor{gray}\multirow{-2}*{24.4} &\cellcolor{gray}\multirow{-2}*{\textbf{79.2}} \\ \hline

\multirow{28}*{Manual-10-3}
&\multirow{7}*{\makecell[l]{PointNet~\cite{qi2017pointnet}}} 
                                &MSP~\cite{hendrycks2016baseline}      &45.4 &63.0 &9.1  &68.1 \\
&                               &MaxLogits~\cite{hendrycks2019scaling} &45.4 &63.0 &8.5  &65.7 \\
&                               &REAL~\cite{cen2022open} &43.2 &57.9 &9.9 &71.6 \\
&                               &UPS+MSP        &44.1 &58.5 &10.0 &71.7 \\
&                               &UPS+MaxLogits  &44.1 &56.2 &10.4 &72.7 \\
&                               &\cellcolor{gray}  &\cellcolor{gray} &\cellcolor{gray} &\cellcolor{gray} &\cellcolor{gray} \\
&                               &\cellcolor{gray}\multirow{-2}*{\makecell[l]{PointCaM \\ (UPS+UPE, Ours)}} 
&\cellcolor{gray}\multirow{-2}*{43.8} &\cellcolor{gray}\multirow{-2}*{42.0} &\cellcolor{gray}\multirow{-2}*{19.8} &\cellcolor{gray}\multirow{-2}*{\textbf{82.8}} \\ \cline{2-7}

&\multirow{7}*{\makecell[l]{PointNet++~\cite{qi2017pointnet++}}}
                                &MSP~\cite{hendrycks2016baseline}      &48.7 &93.4 &6.5 &52.6 \\
&                               &MaxLogits~\cite{hendrycks2019scaling} &48.7 &94.9 &6.4 &51.3 \\
&                               &REAL~\cite{cen2022open} &49.6 &93.4 &6.7 &53.4 \\
&                               &UPS+MSP         &49.3 &93.4 &6.6 &53.3 \\
&                               &UPS+MaxLogits   &49.3 &93.4 &6.9 &54.1 \\
&                               &\cellcolor{gray}  &\cellcolor{gray} &\cellcolor{gray} &\cellcolor{gray} &\cellcolor{gray} \\
&                               &\cellcolor{gray}\multirow{-2}*{\makecell[l]{PointCaM \\ (UPS+UPE, Ours)}}
&\cellcolor{gray}\multirow{-2}*{49.5} &\cellcolor{gray}\multirow{-2}*{84.7} &\cellcolor{gray}\multirow{-2}*{7.8} &\cellcolor{gray}\multirow{-2}*{\textbf{60.6}} \\ \cline{2-7}

&\multirow{7}*{\makecell[l]{PointTransformer~\cite{zhao2021point}}} 
                                &MSP~\cite{hendrycks2016baseline}       &59.7 &100.0 &17.0 &52.8 \\
&                               &MaxLogits~\cite{hendrycks2019scaling}  &59.7 &99.5	 &17.3 &53.0 \\
&                               &REAL~\cite{cen2022open} &61.3 &99.9 &19.5 &58.9 \\
&                               &UPS+MSP         &60.0 &99.8 &19.5 &59.0 \\
&                               &UPS+MaxLogits   &60.0 &93.9 &23.4 &59.8 \\
&                               &\cellcolor{gray}  &\cellcolor{gray} &\cellcolor{gray} &\cellcolor{gray} &\cellcolor{gray} \\
&                               &\cellcolor{gray}\multirow{-2}*{\makecell[l]{PointCaM \\ (UPS+UPE, Ours)}} 
&\cellcolor{gray}\multirow{-2}*{61.6} &\cellcolor{gray}\multirow{-2}*{74.1} &\cellcolor{gray}\multirow{-2}*{26.5} &\cellcolor{gray}\multirow{-2}*{\textbf{70.7}} \\ \cline{2-7}

&\multirow{6}*{\makecell[l]{PointTransformerV3~\cite{wu2024point}}}
                                &MSP~\cite{hendrycks2016baseline}       &58.1 &100.0 &17.6 &54.7 \\

&                               &MaxLogits~\cite{hendrycks2019scaling}  &58.1 &95.2	&19.8 &56.4 \\
&                               &UPS+MSP         &57.3 &99.9 &28.8 &71.2 \\
&                               &UPS+MaxLogits   &57.3 &83.3 &31.8 &72.0 \\
&                               &\cellcolor{gray}  &\cellcolor{gray} &\cellcolor{gray} &\cellcolor{gray} &\cellcolor{gray} \\
&                               &\cellcolor{gray}\multirow{-2}*{\makecell[l]{PointCaM \\ (UPS+UPE, Ours)}}
&\cellcolor{gray}\multirow{-2}*{57.9} &\cellcolor{gray}\multirow{-2}*{65.2} &\cellcolor{gray}\multirow{-2}*{34.7} &\cellcolor{gray}\multirow{-2}*{\textbf{72.4}} \\ \hline

\end{tabular}}
\caption{\textbf{Open-set Point Cloud Semantic Segmentation} on S3DIS\textit{-Split} with PointNet, PointNet++ and PointTransformers. FPR (95$\%$ TPR), pixel-level AUPR, and AUROC in $\%$ are given. $\beta_{max}$ and $\alpha$ are set to 0.6 and 5.0, respectively.}
\label{table:seg1}
\end{center}
\vspace{-20pt}
\end{table*}

In open-set point cloud semantic segmentation, we establish the selection ratio range $[\beta_{min}, \beta_{max}]$ of UPS at $[0.0, 0.6]$. Unlike classification on the sample level, we set $\beta_{min}$ as $0.0$ in semantic segmentation to facilitate the representation of unknown points at various sizes on the pixel level. The selection ratio range for open-set point cloud classification is set at $[0.4, 0.6]$. We set $\beta_{min}$ and $\beta_{max}$ values around $0.5$ to ensure that the mixed unknown samples vary significantly from the original point cloud samples. 
The hyperparameter $\alpha$, which determines the effect of the UPE module, is set to $5.0$ and $1.0$ for segmentation and classification tasks, respectively.

\begin{table*}[h]
\begin{center}

\resizebox{.99\textwidth}{!}{
\begin{tabular}{l|l|cc|ccc}

\hline
& &\multicolumn{2}{c|}{\textbf{Closed-Set Metric}} &\multicolumn{3}{c}{\textbf{Open-set Metric}}
\\ \hline

\textbf{Backbone} &\textbf{Open-set Method} &\textbf{Accuracy (sample)} &\textbf{Accuracy (class)} &\textbf{FPR (95\% TPR) $\downarrow$} &\textbf{Detection Error $\downarrow$} &\textbf{AUROC $\uparrow$} 
\\ \hline

\multirow{5}*{\makecell[l]{PointNet~\cite{qi2017pointnet}}} 
                               &MSP~\cite{hendrycks2016baseline} &96.1 &94.3 &49.9 &17.6 &89.0 \\
                               &MLS~\cite{vaze2021open}  &96.1 &94.3 &50.7 &16.9 &88.8 \\
                               &USS+MSP &96.1 &93.8 &47.4 &17.1 &\textbf{90.2} \\
                               &\cellcolor{gray} &\cellcolor{gray} &\cellcolor{gray} &\cellcolor{gray} &\cellcolor{gray} &\cellcolor{gray}\\
                               &\cellcolor{gray}\multirow{-2}*{\makecell[l]{PointCaM \\ (USS+USE, Ours)}} 
                               &\cellcolor{gray}\multirow{-2}*{96.0} &\cellcolor{gray}\multirow{-2}*{94.1} &\cellcolor{gray}\multirow{-2}*{43.7} &\cellcolor{gray}\multirow{-2}*{17.1} &\cellcolor{gray}\multirow{-2}*{\textbf{90.2}}
                               \\ \hline

\multirow{5}*{\makecell[l]{PointNet++~\cite{qi2017pointnet++}}}
                               &MSP~\cite{hendrycks2016baseline} &97.8 &96.3 &33.5 &14.8 &85.9 \\
                               &MLS~\cite{vaze2021open}   &97.8 &96.3 &35.4 &14.5 &86.2 \\
                               &USS+MSP &97.7 &96.2 &36.4 &15.2 &85.9 \\
                               &\cellcolor{gray} &\cellcolor{gray} &\cellcolor{gray} &\cellcolor{gray} &\cellcolor{gray} &\cellcolor{gray}\\                               
                               &\cellcolor{gray}\multirow{-2}*{\makecell[l]{PointCaM \\ (USS+USE, Ours)}} 
                               &\cellcolor{gray}\multirow{-2}*{97.7} &\cellcolor{gray}\multirow{-2}*{96.3} &\cellcolor{gray}\multirow{-2}*{35.4} &\cellcolor{gray}\multirow{-2}*{15.6} &\cellcolor{gray}\multirow{-2}*{\textbf{86.4}} 
                               \\ \hline

\multirow{5}*{\makecell[l]{DGCNN~\cite{wang2019dynamic}}}
                               &MSP~\cite{hendrycks2016baseline} &97.5 &95.9 &35.9 &13.9 &91.7 \\
                               &MLS~\cite{vaze2021open}   &97.5 &95.9 &34.6 &13.0 &91.7 \\
                               &USS+MSP &97.4 &96.0 &35.1 &13.6 &91.9 \\
                               &\cellcolor{gray} &\cellcolor{gray} &\cellcolor{gray} &\cellcolor{gray} &\cellcolor{gray} &\cellcolor{gray}\\
                               &\cellcolor{gray}\multirow{-2}*{\makecell[l]{PointCaM \\ (USS+USE, Ours)}} 
                               &\cellcolor{gray}\multirow{-2}*{97.7} &\cellcolor{gray}\multirow{-2}*{96.3} &\cellcolor{gray}\multirow{-2}*{35.1} &\cellcolor{gray}\multirow{-2}*{13.7} &\cellcolor{gray}\multirow{-2}*{\textbf{92.5}}
                               \\ \hline

\multirow{5}*{\makecell[l]{AdaptConv~\cite{zhou2021adaptive}}}
                               &MSP~\cite{hendrycks2016baseline} &97.8 &96.3 &33.0 &14.6 &90.6 \\
                               &MLS~\cite{vaze2021open}    &97.8 &96.3 &34.7 &13.5 &89.6 \\
                               &USS+MSP &97.9 &96.4 &33.1 &14.4 &91.8 \\
                               &\cellcolor{gray} &\cellcolor{gray} &\cellcolor{gray} &\cellcolor{gray} &\cellcolor{gray} &\cellcolor{gray}\\
                               &\cellcolor{gray}\multirow{-2}*{\makecell[l]{PointCaM \\ (USS+USE, Ours)}} 
                               &\cellcolor{gray}\multirow{-2}*{97.9} &\cellcolor{gray}\multirow{-2}*{96.6} &\cellcolor{gray}\multirow{-2}*{33.6} &\cellcolor{gray}\multirow{-2}*{14.2} &\cellcolor{gray}\multirow{-2}*{\textbf{91.9}}
                               \\ \hline

\end{tabular}}
\caption{\textbf{Open-set Point Cloud Classification} on ModelNet40\textit{-Split}. The closed-set metrics, including classification accuracies over sample/class, and open-set metrics, including FPR (95$\%$ TPR), detection error, and image-level AUROC in $\%$, are given. ModelNet40\textit{-Split2} is taken as the training dataset, while ModelNet40\textit{-Split2} plus \textit{-Split1} is the evaluation dataset.}
\label{table:cls-model-split21}
\end{center}
\vspace{-20pt}
\end{table*}

\begin{table*}[h]
\begin{center}

\resizebox{.99\textwidth}{!}{
\begin{tabular}{l|l|cc|ccc}

\hline
& &\multicolumn{2}{c|}{\textbf{Closed-Set Metric}} &\multicolumn{3}{c}{\textbf{Open-set Metric}}
\\ \hline

\textbf{Backbone} &\textbf{Open-set Method} &\textbf{Accuracy (sample)} &\textbf{Accuracy (class)} &\textbf{FPR (95\% TPR) $\downarrow$} &\textbf{Detection Error $\downarrow$} &\textbf{AUROC $\uparrow$} 
\\ \hline

\multirow{5}*{\makecell[l]{PointNet~\cite{qi2017pointnet}}} 
                               &MSP~\cite{hendrycks2016baseline} &80.2 &75.7 &86.8 &39.7 &63.6 \\
                               &MLS~\cite{vaze2021open}   &80.2 &75.7 &89.9 &42.2 &60.4 \\
                               &USS+MSP &79.3 &75.5 &90.5 &40.5 &63.3 \\
                               &\cellcolor{gray} &\cellcolor{gray} &\cellcolor{gray} &\cellcolor{gray} &\cellcolor{gray} &\cellcolor{gray}\\
                               &\cellcolor{gray}\multirow{-2}*{\makecell[l]{PointCaM \\ (USS+USE, Ours)}} 
                               &\cellcolor{gray}\multirow{-2}*{79.6} &\cellcolor{gray}\multirow{-2}*{75.6} &\cellcolor{gray}\multirow{-2}*{86.0} &\cellcolor{gray}\multirow{-2}*{38.6} &\cellcolor{gray}\multirow{-2}*{\textbf{65.5}} \\ \hline

\multirow{5}*{\makecell[l]{PointNet++~\cite{qi2017pointnet++}}}
                               &MSP~\cite{hendrycks2016baseline} &88.2 &87.1 &80.6 &30.6	&72.2 \\
                               &MLS~\cite{vaze2021open}  &88.2 &87.1 &80.2 &32.3 &71.4 \\
                               &{USS}+MSP &88.8 &87.8 &81.3 &30.9 &72.2 \\
                               &\cellcolor{gray} &\cellcolor{gray} &\cellcolor{gray} &\cellcolor{gray} &\cellcolor{gray} &\cellcolor{gray}\\
                               &\cellcolor{gray}\multirow{-2}*{\makecell[l]{PointCaM \\ (USS+USE, Ours)}} 
                               &\cellcolor{gray}\multirow{-2}*{88.1} &\cellcolor{gray}\multirow{-2}*{86.8} &\cellcolor{gray}\multirow{-2}*{79.6} &\cellcolor{gray}\multirow{-2}*{30.2} &\cellcolor{gray}\multirow{-2}*{\textbf{73.1}}
                               \\ \hline

\multirow{5}*{\makecell[l]{DGCNN~\cite{wang2019dynamic}}}
                               &MSP~\cite{hendrycks2016baseline} &88.0 &86.2 &84.3 &29.9 &75.3 \\
                               &MLS~\cite{vaze2021open}      &88.0 &86.2 &79.3 &30.3 &75.5 \\
                               &USS+MSP &88.2 &86.2 &83.7 &29.4 &75.5 \\
                               &\cellcolor{gray} &\cellcolor{gray} &\cellcolor{gray} &\cellcolor{gray} &\cellcolor{gray} &\cellcolor{gray}\\
                               &\cellcolor{gray}\multirow{-2}*{\makecell[l]{PointCaM \\ (USS+USE, Ours)}} 
                               &\cellcolor{gray}\multirow{-2}*{87.6} &\cellcolor{gray}\multirow{-2}*{85.9} &\cellcolor{gray}\multirow{-2}*{82.9} &\cellcolor{gray}\multirow{-2}*{29.7} &\cellcolor{gray}\multirow{-2}*{\textbf{76.0}}
                               \\ \hline

\multirow{5}*{\makecell[l]{AdaptConv~\cite{zhou2021adaptive}}}
                               &MSP~\cite{hendrycks2016baseline} &87.5 &85.5 &86.7 &31.1 &73.4 \\                        
                               &MLS~\cite{vaze2021open} &87.5 &85.5 &85.0 &30.5 &{\textbf{74.4}} \\
                               &USS+MSP &86.8 &84.6 &84.6 &30.4 &74.1 \\
                               &\cellcolor{gray} &\cellcolor{gray} &\cellcolor{gray} &\cellcolor{gray} &\cellcolor{gray} &\cellcolor{gray}\\
                               &\cellcolor{gray}\multirow{-2}*{\makecell[l]{PointCaM \\ (USS+USE, Ours)}} 
                               &\cellcolor{gray}\multirow{-2}*{87.1} &\cellcolor{gray}\multirow{-2}*{85.5} &\cellcolor{gray}\multirow{-2}*{84.1} &\cellcolor{gray}\multirow{-2}*{30.1} &\cellcolor{gray}\multirow{-2}*{\textbf{74.4}}
                               \\ \hline

\end{tabular}}
\caption{\textbf{Open-set Point Cloud Classification} on ScanObjectNN\textit{-Split}. The closed-set metrics, including classification accuracies over sample/class, and open-set metrics, including FPR (95$\%$ TPR), detection error, and image-level AUROC in $\%$, are given. ScanObjectNN\textit{-Split2} is taken as the training dataset, while ScanObjectNN\textit{-Split2} plus \textit{-Split1} is the evaluation dataset.}
\label{table:cls-scan-split21}
\end{center}
\vspace{-20pt}
\end{table*}

\subsection{Open-set 3D Semantic Segmentation}
We first conduct open-set experiments on point cloud semantic segmentation, where the model intends to compute the unknown-class scores $\mathcal{X}$ for all testing points. Here, we test our approaches using three popular backbones: the classical PointNet~\cite{qi2017pointnet}, PointNet++~\cite{qi2017pointnet++}, and the advanced PointTransformers~\cite{zhao2021point, wu2024point}.

For most cases in Tab.~\ref{table:seg1}, we observe that the UPS module clearly outperforms the MSP and MaxLogits baselines, as the points simulated by the UPS contribute to modeling the unknown point sets during training. Furthermore, the designed UPE module also significantly improves the open-set metrics. For instance, on the ``Manual-10-3'' split with PointNet, UPS+MSP and UPS+MaxLogits outperform basic MSP and MaxLogits by $3.6$\% and $7.0$\%, respectively, while a full PointCaM (UPS+UPE) mechanism achieves the best performance of $82.8$\%. When testing with PointTransformer, our PointCaM shows significant advantages over the basic MSP/MaxLogits and their incorporation with UPS. For example, on the ``Manual-10-3'' of S3DIS\textit{-Split}, we see PointCaM (UPS+UPE)'s $\sim18$\% and $\sim12$\% of AUROC improvements over MSP and UPS+MSP. Surprisingly, we notice that the network's closed-set accuracies (mIoU) increase from $66.5$/$59.7$\% to $67.1$\%/$61.6$\% with the help of the PointCaM. 
MLUC~\cite{cen2021open} outperforms MSP and MaxLogits with PointNet or PointNet++, but it achieves limited performance improvements using PointTransformer.

In addition to the quantitative results above, we have the following observations: (1) Though PointNet++ and PointTransformer show superior performance in closed-set metrics, they have inferior open-set ability. It indicates that, in 3D semantic segmentation, complicated neural networks make overconfident errors and assign high-confidence predictions to unknown inputs. (2) For open-set problems, two key issues are simulating the negative objects and exploring the spatial/semantic differences between positive and negative data. On the one hand, our UPS and UPE modules are designed to resolve each issue separately; on the other hand, since the UPE module relies on the UPS's simulated {objects} to learn the discrepancies between unknown and known data, the proposed two modules also complement each other for accurate 3D open-set {learning}. 
(3) Tab.~\ref{table:seg1} shows that for PointNet on ``Manual-12-1'', PointCaM does not outperform the baselines in AUPR. While this phenomenon is also observed in other works~\cite{hendrycks2019scaling,cen2022open}, where a lower AUPR accompanies a higher AUROC, the underlying reason still requires further investigation.

\begin{figure*}[t]
\begin{center}
\includegraphics[width=1.0\textwidth]{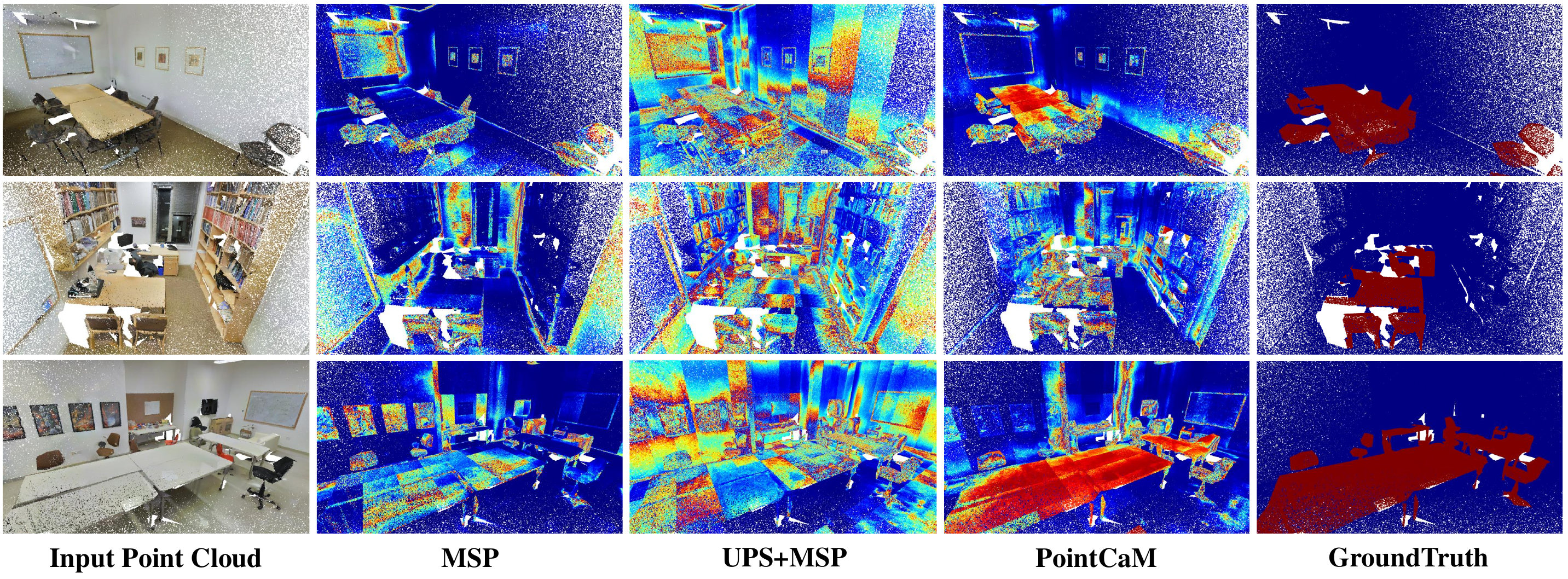}
\caption{{\textbf{The visualization examples: open-set point cloud semantic segmentation} of PointNet on S3DIS\textit{-Split} with the ``Manual-10-3'' split. Under the ``Manual-10-3'' split, ``table,'' ``chair'' and ``sofa'' are taken as unknown classes. We visualize heatmaps of estimated unknown-class scores using different open-set methods, \ie, MSP~\cite{hendrycks2016baseline}, UPS+MSP, and our PointCaM (UPS+UPE). The \textcolor{blue}{``blue''} and \textcolor{red}{``red''} points represent points from known and unknown classes.}}
\label{fig:vis}
\end{center}
\vspace{-10pt}
\end{figure*}

Some visualization examples are given in Figs.~\ref{fig:vis} and \ref{fig:distribution}. 
{Fig.~\ref{fig:vis} visualizes the predictions from open-set point cloud learning models. As we can observe in the second-column subfigures of Fig.~\ref{fig:vis}, the baseline MSP has the essential ability to recognize unknown objects. By using UPS, the model is endowed with superior perception ability for unknown points. For example, in the first-row figure, UPS+MSP identifies more parts of the ``unknown table'' than MSP. However, the UPS module also causes a side effect on the models' recognition confidence in known classes. To address this issue, the UPE module exploits the point cloud's feature context by adaptively fusing the multi-scale feature maps. With the help of both the UPS and UPE modules, our proposed PointCaM mechanism well distinguishes the known and unknown points, as shown in the fourth-column figures.}
The example in Fig.~\ref{fig:distribution} showcases the density distributions of the scores for unknown-class samples. From the figure, it is evident that the separation between known and unknown points becomes more distinct when UPS is employed. This demonstrates that UPS can simulate out-of-distribution points for the network, resulting in increased network sensitivity in identifying unknown points.

\begin{figure*}[t]
\begin{center}
\includegraphics[width=1.0\textwidth]{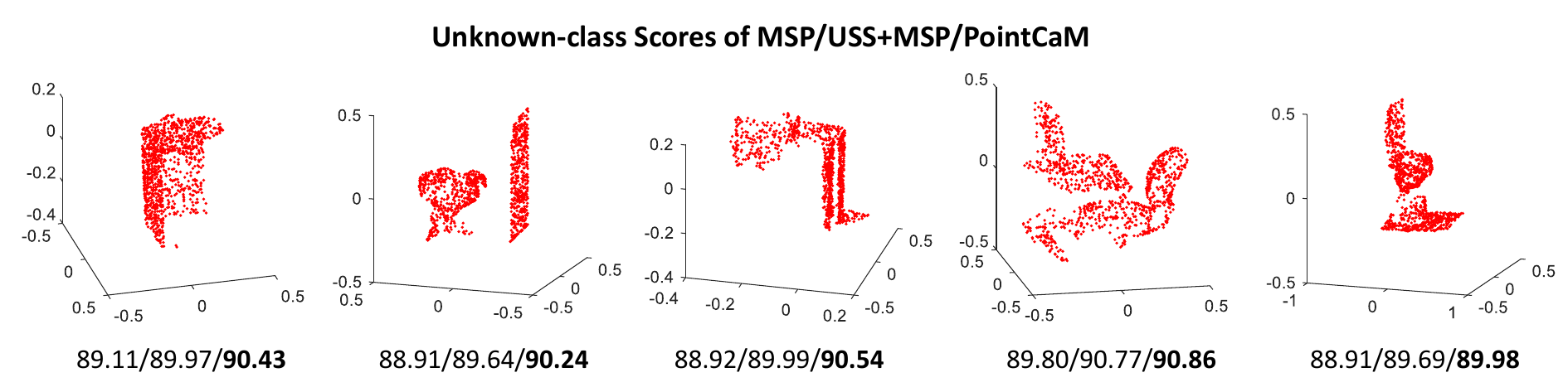}
\caption{{\textbf{The visualization examples: open-set point cloud classification} of AdaptConv on ScanObjectNN\textit{-Split2}. Five unknown point cloud samples are visualized. We provide estimated unknown-class scores using different open-set methods, \ie, MSP~\cite{hendrycks2016baseline}, USS+MSP, and our PointCaM (USS+USE).}}
\label{fig:vis-cls}
\end{center}
\vspace{-20pt}
\end{figure*}

\subsection{Open-set 3D Classification}
Moreover, we transfer our methods to open-set point cloud classification experiments. In addition to PointNet and PointNet++, we utilize DGCNN~\cite{wang2019dynamic} and AdaptConv~\cite{zhou2021adaptive} as testing backbones.

The quantitative results of open-set point cloud classification are reported in Tabs.~\ref{table:cls-model-split21} and \ref{table:cls-scan-split21}, which show that the proposed methods consistently outperform the baseline method MSP in AUROC.
For instance, compared to MSP and USS+MSP, the PointCaM (USS+USE) mechanism is marginally superior in AUROC on two datasets. It is worth noting that, in the ScanObjectNN\textit{-Split} results of Tab.~\ref{table:cls-scan-split21}, the proposed USS+MSP and PointCaM (USS+USE) improve the open-set performance at a small cost to the closed-set accuracy. 
In addition, from the above tables, we find that the enhanced open-set performance by USE (or UPE) is larger than USS (or UPS) in most cases.
One visualization has been provided in Fig.~\ref{fig:vis-cls}. Fig.~\ref{fig:vis-cls} visualizes five samples from unknown classes, where the unknown scores of different approaches are computed and compared. As shown in the figure, the proposed USS and PointCaM give more accurate predictions than MSP.
More experimental results and practical details are provided in the supplementary material.

\subsection{Ablation Study}
\label{sec:abl}

\begin{figure*}[t]
\begin{center}
\includegraphics[width=0.8\textwidth]{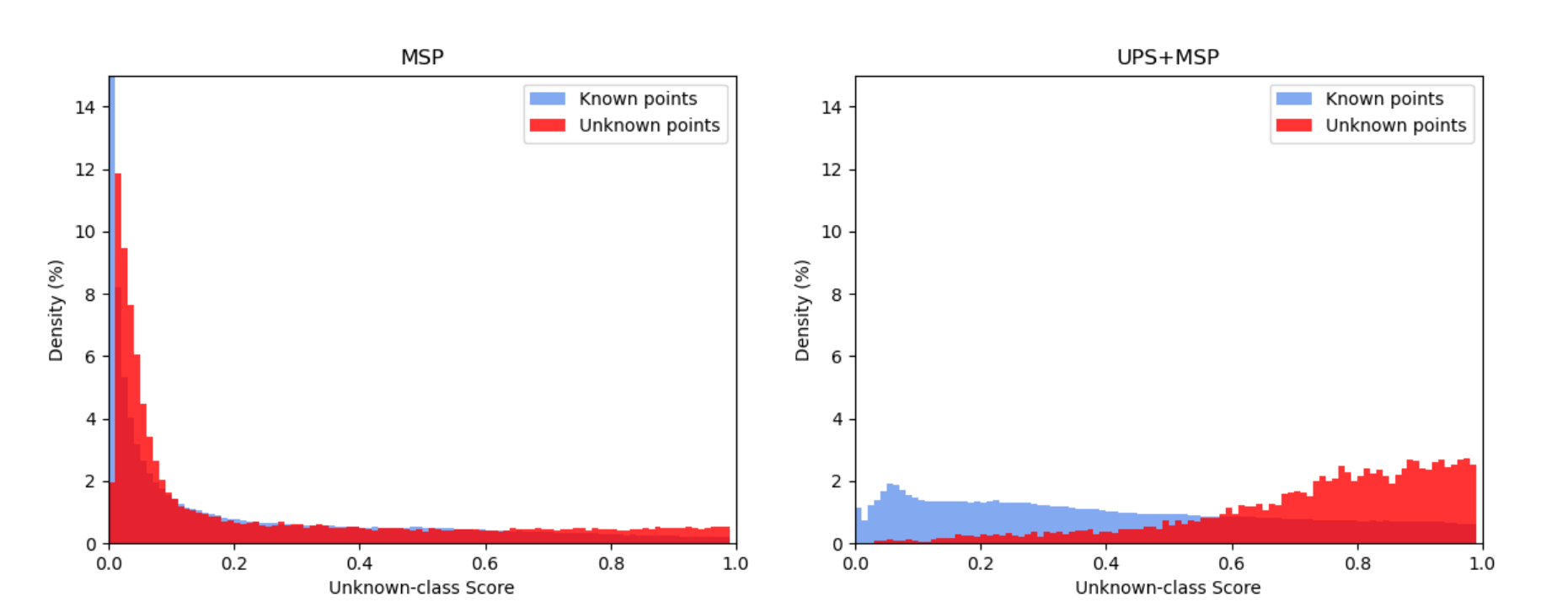}
\caption{{\textbf{The visualization examples: open-set point cloud semantic segmentation} of PointNet on S3DIS\textit{-Split} with the ``Manual-10-3'' split. We visualize the density distributions of the unknown-class score of MSP and UPS+MSP.}}
\label{fig:distribution}
\end{center}
\vspace{-10pt}
\end{figure*}

\subsubsection{Unknown point generator}
Here, we test multiple approaches to simulating unknown points: single rotation within different angle ranges, single translation within different ranges of the entire point cloud, scaling, and Gaussian noise. Simple rotation and translation can be considered special cases of our UPS when the translation vector is zero or the rotation matrix is the identity. 
{Results in Tab.~\ref{table:generator} show that UPS can provide an effective baseline for open-set point cloud learning. Compared to other methods, UPS, which can generate objects of natural and diverse appearances, provides better simulations of the out-of-distribution point cloud objects.}

\begin{table}[h]
\begin{center}
\resizebox{.9\textwidth}{!}{
\begin{tabular}{l|l|l|c}
\hline
\textbf{Data Split} &\textbf{Backbone} &\textbf{Open-set Method} &\textbf{AUROC $\uparrow$} 
\\ \hline

\multirow{10}*{Manual-10-3}
&\multirow{10}*{\makecell[l]{PointNet~\cite{qi2017pointnet}}}
                                &MSP                                  &68.1 \\ \cline{3-4}
                                
&                               &{MSP+single rotation (0.5$\pi$)}  & 69.9 \\
&                               &{MSP+single rotation ($\pi$)}     & 70.4 \\
&                               &MSP+single rotation (2$\pi$)      & 70.9 \\ \cline{3-4}

&                               &{MSP+single translation (0.25 range)}  &63.9 \\
&                               &{MSP+single translation (0.5 range)}   &69.5 \\
&                               &MSP+single translation (whole range)   &70.8 \\ \cline{3-4}

&                               &{MSP+scaling}        &{67.3} \\ \cline{3-4}
&                               &{MSP+Gaussian noise} &{58.1} \\ \cline{3-4}
&                               &MSP+UPS                              &\textbf{71.7} \\
\hline

\end{tabular}}
\caption{{\textbf{The ablation study:} Unknown point generator. PointNet is performed in Open-set Point Cloud Semantic Segmentation on S3DIS\textit{-Split}. The metric AUROC in $\%$ is given.}}
\label{table:generator}
\end{center}
\vspace{-20pt}
\end{table}

\subsubsection{Hyperparameters}
There are two main hyperparameters in the PointCaM mechanism: the maximum selection ratio $\beta_{max}$ in the UPS module and the coefficient of the MSE loss $\alpha$ in the UPE module.
The values for both parameters are tuned and selected using the grid search.
From Tabs.~\ref{table:beta} and \ref{table:alpha}, models show sensitivity to these two parameters. The results in Tab.~\ref{table:beta} show that with growing $\beta_{max}$, the open-set recognition ability of the model increases. Given the testing performances of $\alpha$ shown in Tab.~\ref{table:alpha}, we set the value $\alpha = 5.0$ for our experiments using the PointCaM mechanism.
{From Tab.~\ref{table:alpha}, the larger value of $\alpha$ ($>5.0$) causes the performance drop since UPE might be overly adapted to simulated out-of-distribution points rather than genuine unknown objects.}

\begin{table}[h]
\begin{center}
\resizebox{.9\textwidth}{!}{
\begin{tabular}{l|l|l|c|ccc}
\hline
\textbf{Data Split} &\textbf{Backbone} &\textbf{Open-set Method} &\textbf{$\beta_{max}$} &\textbf{AUROC $\uparrow$} 
\\ \hline

\multirow{13}*{Manual-12-1}
&\multirow{13}*{\makecell[l]{PointTransformer~\cite{zhao2021point}}}  
                                &MSP &- &59.0  \\ \cline{3-5}

&                               &UPS+MSP        &\multirow{2}*{0.2}  &59.4\\
&                               &UPS+MaxLogits  &                    &62.3\\ \cline{3-5}

&                               &UPS+MSP        &\multirow{2}*{{0.3}}  &62.8 \\
&                               &UPS+MaxLogits  &                      &64.7 \\ \cline{3-5}

&                               &UPS+MSP        &\multirow{2}*{0.4}  &61.6\\
&                               &UPS+MaxLogits  &                    &64.6\\ \cline{3-5}

&                               &UPS+MSP        &\multirow{2}*{{0.5}}  &63.5 \\
&                               &UPS+MaxLogits  &                      &65.1 \\ \cline{3-5}

&                               &UPS+MSP        &\multirow{2}*{0.6}  &\textbf{66.0} \\
&                               &UPS+MaxLogits  &                    &\textbf{65.4} \\ \cline{3-5}

&                               &UPS+MSP        &\multirow{2}*{{0.7}}  &65.5 \\
&                               &UPS+MaxLogits  &                      &65.0 \\ \hline

\end{tabular}}
\caption{{\textbf{The ablation study: $\beta_{max}$.} PointTransformer is performed in Open-set Point Cloud Semantic Segmentation on S3DIS\textit{-Split}. The metric AUROC in $\%$ is given. Ranging $\beta_{max}$ does not affect outperforming the baseline MSP.}}
\label{table:beta}
\end{center}
\vspace{-20pt}
\end{table}

\begin{table}[h]
\begin{center}
\resizebox{.9\textwidth}{!}{
\begin{tabular}{l|l|c|c|c}
\hline
\textbf{Data Split} &\textbf{Backbone} &\textbf{Open-set Method} &\textbf{$\alpha$} &\textbf{AUROC $\uparrow$} 
\\ \hline

\multirow{9}*{Manual-10-3}
&\multirow{9}*{\makecell[l]{PointNet++~\cite{qi2017pointnet++}}}
                                &MSP      &-    &52.6 \\ \cline{3-5}
&                               &\multirow{8}*{PointCaM}    &0.2  &56.8 \\
&                               &                           &1.0  &59.5 \\
&                               &                           &2.5  &58.6 \\
&                               &                           &5.0  &\textbf{60.6} \\
&                               &                           &7.5  &57.0 \\
&                               &                           &10.0 &58.3 \\
&                               &                           &12.5 &56.6 \\
&                               &                           &15.0 &55.5 \\
\hline
\end{tabular}}
\caption{{\textbf{The ablation study: $\alpha$.} PointNet++ is performed in Open-set Point Cloud Semantic Segmentation on S3DIS\textit{-Split}. The metric AUROC in $\%$ is given. Ranging $\alpha$ does not affect outperforming the baseline MSP.}}
\label{table:alpha}
\end{center}
\vspace{-10pt}
\end{table}

\subsubsection{Fusion layers in UPE}
In addition to UPS, which simulates unknown points, we propose UPE to enhance open-set performance by leveraging feature context in point clouds. In practice, UPE is designed to fuse the feature context from the encoders at different levels. We present an ablation study for UPE in Tab.~\ref {table:upe}, demonstrating that its performance improves when different levels of feature context are fused.

\begin{table}[h]
\begin{center}
\resizebox{.95\textwidth}{!}{
\begin{tabular}{l|l|c|l|c}
\hline
\textbf{Data Split} &\textbf{Backbone} &\textbf{Open-set Method} &\textbf{Fusion Layer} &\textbf{AUROC $\uparrow$} 
\\ \hline

\multirow{7}*{Manual-10-3}
&\multirow{7}*{\makecell[l]{PointNet~\cite{qi2017pointnet}}}
                                &\multirow{7}*{PointCaM}    &-                  &68.1 \\ \cline{4-5}
&                               &                           &1st layer          &78.8 \\ \cline{4-5}
&                               &                           &1st+2nd layers     &81.2 \\ \cline{4-5}
&                               &                           &\multirow{2}*{\makecell[l]{{1st+2nd+3rd layers}\\{(w/o point-guided fusion)}}}  &\multirow{2}*{{79.9}} \\
&                               &                           && \\ \cline{4-5}
&                               &                           &\multirow{2}*{\makecell[l]{1st+2nd+3rd layers\\(with point-guided fusion)}}  &\multirow{2}*{\textbf{82.8}} \\
&                               &                           && \\
\hline
\end{tabular}}
\caption{\textbf{The ablation study:} Fusion layers in UPE. PointNet++ is performed in Open-set Point Cloud Semantic Segmentation on S3DIS\textit{-Split}. The metric AUROC in $\%$ is given.}
\label{table:upe}
\end{center}
\vspace{-20pt}
\end{table}

\subsection{Discussion}
\subsubsection{Number of unknown classes}
The results on S3DIS\textit{-Split} (See Tab.~\ref{table:seg1}) imply the negative impacts of unknown data on closed-set accuracy: increasing the number of unknown classes causes a worse closed-set accuracy. Thus, the mIoU values of the ``Manual-10-3'' split are lower than the ones reported in the ``Manual-12-1'' split. This phenomenon aligns with the statements in~\cite{ma2018towards, dhamija2020overlooked}.

\subsubsection{2D open-set methods}
In Tab.~\ref{table:seg1}, the incremental improvements produced by MSP~\cite{hendrycks2016baseline} and MaxLogits~\cite{hendrycks2019scaling} demonstrate that the 2D-driven methods also work for 3D point clouds. 
However, the performances in 2D and 3D cases are not consistent: \eg, MaxLogits usually performs better than MSP in 2D tasks as~\cite{hendrycks2019scaling} verifies. In contrast, in our 3D experiments, the performance of MaxLogits is inferior to that of MSP (See Tab.~\ref {table:seg1}). Similar observations showing possible incompatibility between 2D and 3D-driven open-set methods can be found in~\cite{cen2022open}.
Due to the unordered and unstructured characteristics of point clouds, directly applying 2D methods to 3D tasks may not be preferable. Thus, 3D-specific open-set methods such as our proposed PointCaM are worth further exploration.

\subsubsection{Training time}
In Tab.~\ref{table:complex}, we compare the training time and model size of various approaches with PointNet on ``Manual-12-1''. The table shows that although there is a slight increase in model complexity, the substantial improvements in the open-set metric of our PointCaM model justify the extra complexity. UPS+MSP uses less training time, as at some iteration, a portion of the multi-class data is manipulated as unknown data of only one class, thereby reducing the complexity of the model training.

\begin{table*}[t]
\begin{center}
\resizebox{1.0\textwidth}{!}{
\begin{tabular}{l|l|l|cc|c}

\hline
&&&\multicolumn{2}{c|}{\textbf{{Model Complexity}}} &\textbf{Open-set Metric}
\\ \hline

\textbf{Data Split} &\textbf{Backbone} &\textbf{Open-set Method} &\textbf{{Training Time}} &\textbf{{Params}} &\textbf{AUROC $\uparrow$} 
\\ \hline

\multirow{5}*{Manual-10-3}
&\multirow{5}*{\makecell[l]{PointNet~\cite{qi2017pointnet}}} 
                                &MSP~\cite{hendrycks2016baseline}        &{0.23s} &{41481K}  &68.1 \\
&                               &MaxLogits~\cite{hendrycks2019scaling}   &{0.23s} &{41481K}  &65.7 \\
&                               &UPS+MSP          &{0.19s} &{41482K}  &71.7 \\
&                               &UPS+MaxLogits    &{0.19s} &{41482K}  &72.7 \\
&                               &\cellcolor{gray}PointCaM (Ours) &\cellcolor{gray}{0.24s} &\cellcolor{gray}{41501K} &\cellcolor{gray}\textbf{82.8} \\ \hline
\end{tabular}}
\caption{{\textbf{Training time:} PointNet is performed in Open-set Point Cloud Semantic Segmentation on S3DIS\textit{-Split}. The training time for each iteration and model size are provided.}}
\label{table:complex}
\end{center}
\vspace{-10pt}
\end{table*}

\subsubsection{Other benchmarks}
In addition to the indoor scene dataset, we perform experiments on the autonomous driving datasets SemanticKITTI~\cite{behley2019semantickitti}.
The results in Tab.~\ref{table:REAL} demonstrate that PointCaM achieves comparable performance to other methods. 
As seen from the table, PointCaM achieves the highest AUROC. It is noted that even though PointCaM surpasses C3D+APF~\cite{li2023open} in AUROC and mIoU, C3D+APF obtains a much higher AUPR.
We also evaluate our PointCaM on the 3DOS benchmark~\cite{alliegro20223dos} for open-set point cloud classification. The results are presented in Tab.~\ref{table:3DOS}, where PointCaM delivers comparable performance to the best algorithms performed on the 3DOS benchmark.
In this work, we evaluate our approach on datasets such as S3DIS and SemanticKITTI, which collect data from the real world. This demonstrates the potential practical utility of our approach in real-world scenarios.

\begin{table}[h]
\begin{center}
\resizebox{.50\textwidth}{!}{
\begin{tabular}{l|cc}
\hline

Dataset &\multicolumn{2}{c}{SemanticKITTI} \\ \hline
Methods &AUROC $\uparrow$ &mIoU $\uparrow$ \\ \hline

MSP~\cite{hendrycks2016baseline}      &74.0  &58.0  \\
MaxLogits~\cite{hendrycks2019scaling} &70.5  &58.0  \\ 
MC-Dropout~\cite{gal2016dropout}      &74.7  &\textbf{58.0}  \\
REAL~\cite{cen2022open}               &84.9  &57.8  \\
C3D+APF~\cite{li2023open}             &85.6  &57.3  \\
DOSS~\cite{deng2025novel}             &\textbf{87.6}  &57.0  \\ \hline
\rowcolor{gray} {PointCaM (Ours)}            &\textbf{86.2}  &\textbf{57.8} \\ \hline

\end{tabular}}
\caption{\textbf{Other benchmarks:} Open-set point cloud semantic segmentation on SemanticKITTI~\cite{behley2019semantickitti}. AUROC and mIoU in $\%$ are given. The best two numbers are in bold. The hyperparameters $\beta$ and $\alpha$ are set to $0.6$ and $150.0$.}
\label{table:REAL}
\end{center}
\vspace{-20pt}
\end{table}

\begin{table}[t]
\begin{center}
\resizebox{.95\textwidth}{!}{
\begin{tabular}{l|cc|cc|cc}
\hline

\multicolumn{7}{c}{Synth to Real Benchmark - DGCNN} \\ \hline
&\multicolumn{2}{c|}{SR 1 (easy)}  &\multicolumn{2}{c|}{SR 2 (hard)}  &\multicolumn{2}{c}{Avg} \\
Method &AUROC $\uparrow$ &FPR95 $\downarrow$ &AUROC $\uparrow$ &FPR95 $\downarrow$ &AUROC $\uparrow$ &FPR95 $\downarrow$ \\ \hline

MSP             &72.2 &91.0 &61.2 &90.3 &66.7 &90.6 \\
MLS             &69.0 &92.2 &62.4 &88.9 &65.7 &90.5 \\ 
ODIN            &69.0 &92.2 &62.4 &89.0 &65.7 &90.6 \\
Energy          &68.8 &92.7 &62.4 &88.9 &65.6 &90.8 \\
GradNorm        &67.0 &93.5 &59.8 &89.4 &63.4 &91.5 \\
ReAct           &68.4 &92.1 &62.8 &88.8 &65.6 &90.5 \\

VAE             &68.6 &77.0 &57.9 &92.3 &63.3 &84.6 \\
NF              &72.5 &81.6 &70.2 &83.0 &71.3 &\textbf{82.3} \\
OE+mixup        &71.1 &89.6 &59.5 &92.0 &65.3 &90.8 \\
ARPL+CS         &71.5 &90.2 &62.8 &89.5 &67.1 &89.8 \\
Cosine proto    &58.6 &90.6 &57.3 &91.3 &57.9 &91.0 \\
CE              &67.5 &87.4 &64.6 &91.0 &66.1 &89.2 \\
SubArcFace      &74.5 &86.7 &68.7 &86.6 &\textbf{71.6} &86.7 \\ \hline
\rowcolor{gray} {PointCaM (Ours)} &{77.04} &{87.52} &{65.23} &{85.59} &{71.1} &{86.6} \\ \hline

\end{tabular}}
\caption{{\textbf{Other benchmarks:}} Open-set point cloud classification on 3DOS (Synthetic to Real Benchmark track)~\cite{alliegro20223dos}. The metrics AUROC and FPR95 in $\%$ are given. {The hyperparameters $\beta$ and $\alpha$ are set as $[0.4, 0.6]$ and $0.1$.}}
\label{table:3DOS}
\end{center}
\vspace{-20pt}
\end{table}
 
\subsubsection{Limitation}
We identify the limitations of our work: (1) Although our empirical results demonstrate the superiority of our PointCaM model, the theoretical underpinnings of our findings require further development to explain our observations better. (2) In some instances, the close-set metrics are marginally inferior to those of the baseline. For example, as illustrated in Tab.~\ref{table:cls-scan-split21}, PointCaM with PointNet backbone attains an \textit{$\mathrm{mIoU}$} of $79.6\%$ on ScanObjectNN\textit{-Split2}, which is lower than $80.2\%$ of MSP. Although this finding is consistent with some prior studies~\cite{chen2021adversarial, cen2022open}, the reasons for this outcome require further investigation. 
(3) As observed in Tabs.~\ref{table:beta} and \ref{table:alpha}, our proposed model appears to be somewhat sensitive to hyperparameters due to the inherent diversities of 3D point cloud data.
 
\subsubsection{Performance drop}
The performance drops in known classes are found in the above tales, although the open-set performances are enhanced. 
Similar phenomena have been observed in prior studies~\cite{chen2021adversarial, cen2022open, li2023open, deng2025novel}. 
In~\cite{li2023open}, the mIoU of the proposed method drops from $58.0\%$ to $57.3\%$ compared to the baseline. 
The mIoU of the proposed method in \cite{deng2025novel} is lower than that of the SOTA methods by $0.3\%$ to $0.8\%$.
We conduct an experiment to investigate the cause of the performance drop, and the results are presented in Tab.~~\ref{table:drop}. From the above tables, we observe that the open-set performance basically follows a similar trend to the closed-set performance. For instance, when AUROC drops from $86.4\%$ to $84.3\%$, the classification accuracy decreases by $0.5\%$ from $97.7\%$ to $97.2\%$. We also test a wider range of values for the augmentation ratio hyperparameter, from $0.01$ to $0.2$, and find that an increase in AUROC results in slightly lower closed-set classification accuracy. Additionally, in this case, the FPR (95\% TPR) and Detection Error exhibit a similar trend to the classification accuracy. 
We infer that as the model's ability to distinguish unknown samples improves (indicated by an increase in AUROC), it becomes more cautious and makes more conservative decisions, \ie, misclassifying some known samples as unknown ones. This hypothesis is somewhat supported by the visualization in Fig.~\ref{fig:distribution}, where a more distinct separation between known and unknown samples results in lower confidence scores for the known samples. 
Based on the discussion above, achieving a trade-off performance on both known and unknown data could be a research direction in the open-set problem towards practical applications.

\begin{table*}[h]
\begin{center}

\resizebox{.89\textwidth}{!}{
\begin{tabular}{l|c|c|ccc}

\hline
& &\multicolumn{1}{c|}{\textbf{Closed-Set Metric}} &\multicolumn{3}{c}{\textbf{Open-set Metric}}
\\ \hline

\textbf{Backbone} &\textbf{Ratio} &\textbf{Accuracy (sample)} &\textbf{FPR (95\% TPR) $\downarrow$} &\textbf{Detection Error $\downarrow$} &\textbf{AUROC $\uparrow$} 
\\ \hline

\multirow{6}*{\makecell[l]{PointNet++~\cite{qi2017pointnet++}}} 
                               &0.01 &{97.9} &{33.4} &{14.7} &{85.9} \\
                               &0.05 &{97.8} &{35.4} &{15.5} &{86.2} \\
                               &0.1  &97.7 &35.4 &15.6 &\textbf{86.4} \\
                               &0.2  &{97.8} &{34.2} &{15.0} &{86.1} \\ \cline{2-6}
                               &0.3  &{97.4} &{42.0} &{16.2} &{84.8} \\ 
                               &0.5  &{97.2} &{43.0} &{17.1} &{84.3} \\ \hline

\end{tabular}}

\vspace{10pt}
\resizebox{.9\textwidth}{!}{
\includegraphics[width=1.0\textwidth]{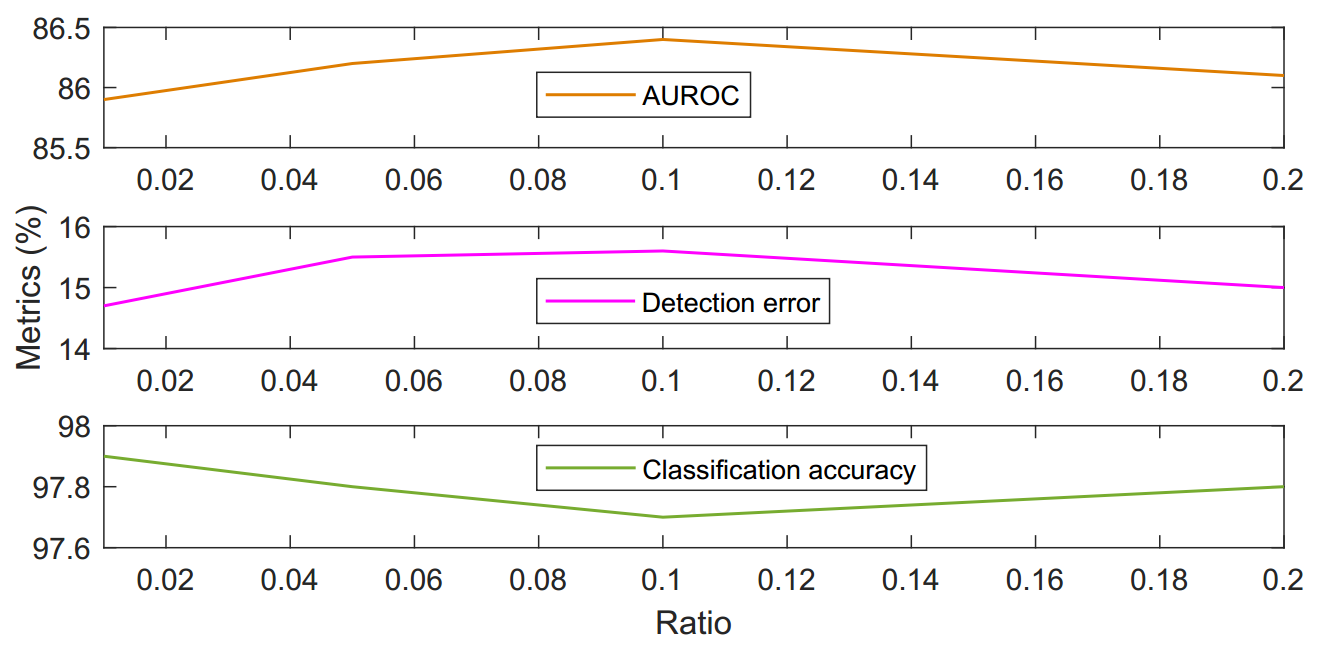}}

\caption{{\textbf{Performance drop:} Open-set point cloud classification of PointNet++ on ModelNet40-Split with different ratios of using samples as unknown ones.}}
\label{table:drop}
\end{center}
\vspace{-20pt}
\end{table*}
\section{Conclusion}\label{Conclusion}
This paper discusses point cloud learning under open-set settings. Two main point cloud visual tasks, semantic segmentation and classification, have been deeply investigated. To address the open-set point cloud problems, we propose a PointCaM mechanism to enhance a model's ability to identify unknown objects. Specifically, the UPS module helps simulate out-of-distribution data to some extent during the training stage. In contrast, the UPE module aims to leverage the fundamental context of feature maps and point cloud data for accurate computation of unknown-class scores. By conducting extensive experiments and comprehensive ablation studies, we verify the effectiveness of the proposed approach in open-set point cloud learning. The decline in the closed-set metric is observed; therefore, one potential future work is to maintain closed-set performance in open-set 3D scenarios. Additionally, the application of 3D foundation models in simulating unknown objects could be explored. Finally, we hope our work could inspire future investigations into open-set settings for more point cloud problems, \eg, point cloud reconstruction, completion, and upsampling.

\bibliographystyle{elsarticle-num}
\bibliography{main.bib}

\end{document}